\documentclass[10pt,twocolumn,letterpaper]{article}

\usepackage[pagenumbers]{cvpr}

\newcommand{\tokenizername}{DeltaTok\xspace}
\newcommand{\modelname}{DeltaWorld\xspace}

\usepackage[protrusion=true,expansion=true]{microtype}

\usepackage{amssymb}
\usepackage{multirow}
\usepackage[table]{xcolor}
\usepackage{tikz,pgfplots}
\usetikzlibrary{arrows.meta,positioning,calc}
\usepackage{fontawesome5}
\newcommand{\frozen}{\textcolor{cyan!80!blue}{\footnotesize\faSnowflake}}
\newcommand{\train}{\textcolor{red!80}{\footnotesize\faFire}}

\pgfplotsset{compat=1.18}
\definecolor{rowgray}{gray}{0.92}

\definecolor{vfmfill}{HTML}{FAD7AC}
\definecolor{vfmstroke}{HTML}{B46504}
\definecolor{tokfill}{HTML}{FFF2CC}
\definecolor{tokstroke}{HTML}{D6B656}
\definecolor{lossfill}{HTML}{C8CCD0}
\definecolor{lossstroke}{HTML}{707880}
\definecolor{vitfill}{HTML}{FFF2CC}
\definecolor{vitstroke}{HTML}{D6B656}
\definecolor{boxfill}{HTML}{FFFFFF}
\definecolor{boxstroke}{HTML}{707880}
\definecolor{arrgray}{HTML}{C8CCD0}
\definecolor{patchA}{HTML}{C8CCD0}
\definecolor{patchB}{HTML}{C8CCD0}
\definecolor{patchZ}{HTML}{5CB85C}
\definecolor{tokenbox}{HTML}{FFFFFF}
\definecolor{crossfill}{HTML}{FFF2CC}
\definecolor{crossstroke}{HTML}{D6B656}
\definecolor{outfill}{HTML}{B0E3E6}
\definecolor{outstroke}{HTML}{0E8088}
\definecolor{bestfill}{HTML}{44DD44}
\definecolor{beststroke}{HTML}{228B22}

\newcommand{\downrightarrow}{\hspace{0pt}%
  \raisebox{1.5pt}{\begin{tikzpicture}[scale=0.4, baseline=(current bounding box.south)]
    \draw[-latex] (0.0, 0.5) -- (0.0, 0.25) -- (0.55, 0.25);
  \end{tikzpicture}} %
}

\newcommand{\tablestep}[1]{\textcolor{gray!90}{(#1)}}

\usepackage{tabularx,array}
\newcolumntype{Y}{>{\raggedright\arraybackslash}X}
\usepackage{multirow}
\definecolor{cvprblue}{rgb}{0.21,0.49,0.74}
\usepackage[pagebackref,breaklinks,colorlinks,allcolors=cvprblue]{hyperref}
\usepackage[accsupp]{axessibility}

\def\confName{CVPR}
\def\confYear{2026}

\title{A Frame is Worth One Token:\\ Efficient Generative World Modeling with Delta Tokens}

\author{
Tommie Kerssies\textsuperscript{1,2,*} \quad
Gabriele Berton\textsuperscript{1,*} \quad
Ju He\textsuperscript{1} \quad
Qihang Yu\textsuperscript{1} \quad
Wufei Ma\textsuperscript{1,3,*} \\
Daan de Geus\textsuperscript{2,**} \quad
Gijs Dubbelman\textsuperscript{2,**} \quad
Liang-Chieh Chen\textsuperscript{1,*}
\\[0.5em]
\textsuperscript{1}Amazon \quad
\textsuperscript{2}Eindhoven University of Technology \quad
\textsuperscript{3}Johns Hopkins University
\\[0.3em]
{\small \textsuperscript{*}Work done while at Amazon. \textsuperscript{**}Equal advising.}
}

\begin{document}
\maketitle
\begin{abstract}
Anticipating diverse future states is a central challenge in video world modeling.
Discriminative world models produce a deterministic prediction that implicitly averages over possible futures, while existing generative world models remain computationally expensive.
Recent work demonstrates that predicting the future in the feature space of a vision foundation model (VFM), rather than a latent space optimized for pixel reconstruction, requires significantly fewer world model parameters.
However, most such approaches remain discriminative.
In this work, we introduce \tokenizername, a tokenizer that encodes the VFM feature difference between consecutive frames into a single continuous ``delta'' token, and \modelname, a generative world model operating on these tokens to efficiently generate diverse plausible futures. Delta tokens reduce video from a three-dimensional spatio-temporal representation to a one-dimensional temporal sequence, for example yielding a $1{,}024\times$ token reduction with $512 \times 512$ frames. This compact representation enables tractable multi-hypothesis training, where many futures are generated in parallel and only the best is supervised. At inference, this leads to diverse predictions in a single forward pass.
Experiments on dense forecasting tasks demonstrate that \modelname forecasts futures that more closely align with real-world outcomes, while having over $35\times$ fewer parameters and using $2{,}000\times$ fewer FLOPs than existing generative world models.
Code \& weights: \href{https://deltatok.github.io}{\texttt{deltatok.github.io}}.
\end{abstract}

\section{Introduction}
The ability to predict future states of the world is essential for autonomous robots and vehicles. A \textit{world model}~\cite{ha2018worldmodels} provides this capability, enabling agents to anticipate upcoming events and plan safe, effective actions. Because the future is inherently uncertain, predictions must account for multiple possible future world states.
In autonomous driving, for instance, anticipating interactions among multiple agents requires reasoning over diverse futures to prevent collisions.

\begin{figure}[!t]
  \centering
  \resizebox{\linewidth}{!}{%
  \begin{tikzpicture}[
      rect_existing/.style={
          draw=vfmstroke,
          fill=vfmfill,
          rounded corners=0.15cm,
          inner sep=0.15cm,
          minimum width=2.6cm,
          minimum height=1.6cm,
          align=center,
          font=\sffamily\scriptsize
      },
      rect_ours/.style={
          draw=tokstroke,
          fill=tokfill,
          rounded corners=0.15cm,
          inner sep=0.15cm,
          minimum width=1.8cm,
          minimum height=0.6cm,
          align=center,
          font=\sffamily\scriptsize
      },
      rect_frame/.style={
          draw=lossstroke,
          fill=white,
          rounded corners=0.08cm,
          inner sep=0.15cm,
          minimum width=1.1cm,
          minimum height=1.1cm,
          align=center
      },
      rect_token_spatial/.style={
          draw=patchA!50!black,
          fill=patchA,
          minimum width=0.22cm,
          minimum height=0.22cm,
          inner sep=0pt,
          rounded corners=1pt
      },
      rect_token_delta/.style={
          draw=patchZ!50!black,
          fill=patchZ,
          minimum width=0.22cm,
          minimum height=0.22cm,
          inner sep=0pt,
          rounded corners=1pt
      }
  ]

  \definecolor{copylast}{HTML}{E69F00}

  \node[rect_existing, anchor=east] (existing_models) at (1.5, 0) {
      Existing\\
      generative\\
      world models
  };
  
  \node[font=\sffamily\scriptsize, anchor=north] at (existing_models.south) {Big model};
  
  \begin{scope}[shift={(3.6, 0)}]
    \foreach \i in {0, 1, 2} {
      \node[rect_frame] (f\i) at (\i*1.3, 0) {};
      \begin{scope}[shift={({(\i*1.3)-0.32}, {-0.32})}]
        \foreach \x in {0, 1, 2} {
            \foreach \y in {0, 1, 2} {
                \node[rect_token_spatial] at (\x*0.32, \y*0.32) {};
            }
        }
      \end{scope}
    }
  \end{scope}
  
  \draw[->, >=stealth, line width=0.8pt, lossstroke] (existing_models.east) -- (3.0, 0)
      node[midway, above, font=\sffamily\scriptsize, text=black] {Many}
      node[midway, below, font=\sffamily\scriptsize, text=black] {forwards};

  \draw[->, >=stealth, line width=0.8pt, lossstroke, rounded corners=3pt]
      (f2.east) -- ++(0.4, 0) -- ++(0, 1.2) --
      (existing_models.north |- 0, 1.2) -- (existing_models.north);

  \node[font=\sffamily\scriptsize, anchor=north] at (4.9, 0 |- existing_models.south) {Many spatial tokens per future};

  \begin{scope}[shift={(0, -2.2)}]
    \node[rect_ours, anchor=center] (deltaworld) at (existing_models.center |- 0, 0) {
        \textbf{\modelname}\\\textbf{(Ours)}
    };
    
    \node[font=\sffamily\scriptsize, anchor=north] (small_model) at (deltaworld.south) {\emph{Small} model};
    
    \foreach \xpos in {3.6, 4.9, 6.2} {
      \node[draw=lossstroke, fill=white, rounded corners=0.08cm, inner sep=0.12cm, minimum width=0.5cm, minimum height=0.5cm] at (\xpos, 0) {};
      \node[rect_token_delta] at (\xpos, 0) {};
    }
    
    \draw[->, >=stealth, line width=0.8pt, lossstroke] (deltaworld.east) -- (3.35, 0)
        node[midway, above, font=\sffamily\scriptsize, text=black] {\emph{One}}
        node[midway, below, font=\sffamily\scriptsize, text=black] {forward};
    
    \node[font=\sffamily\scriptsize, anchor=north] at (4.9, 0 |- deltaworld.south) {\emph{One} delta token per future};
  \end{scope}

  \draw[line width=0.8pt, densely dashed, lossstroke] (current bounding box.west |- 0,-1.45) -- (current bounding box.east |- 0,-1.45);

  \end{tikzpicture}%
  } %
  \caption{\textbf{Outline of \modelname.} Unlike large existing generative world models that require many forward passes and represent each frame with many spatial tokens, our small \modelname generates multiple futures in a single forward pass by using a single \emph{delta token} to encode the difference between consecutive frames.
  }
  \label{fig:teaser}
\end{figure}
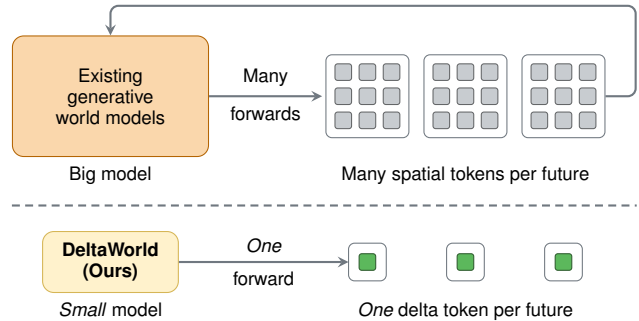

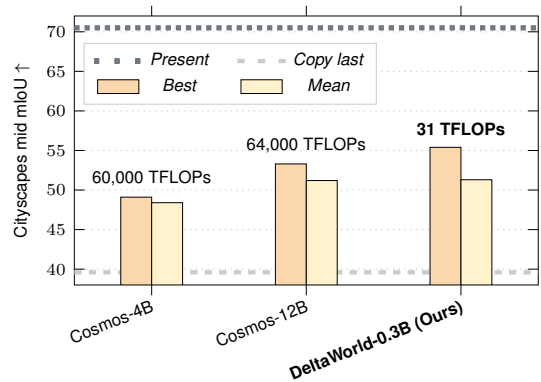
\begin{figure}[t]
  \centering
  \resizebox{0.85\linewidth}{!}{%
  \begin{tikzpicture}

  \definecolor{bestbar}{HTML}{7DAA22}      %
  \definecolor{meanbar}{HTML}{E69F00}      %
  \definecolor{copylast}{HTML}{C8CCD0}
  \definecolor{presentline}{HTML}{707880}

  \begin{axis}[
      ybar=0pt,
      bar width=0.45cm,
      width=\linewidth,
      height=5.5cm,
      enlarge x limits=0.25,
      ymin=38, ymax=72,
      ytick={40, 45, 50, 55, 60, 65, 70},
      ylabel={Cityscapes mid mIoU $\uparrow$},
      ylabel style={font=\sffamily\scriptsize},
      y tick label style={font=\sffamily\scriptsize},
      symbolic x coords={Cosmos-4B, Cosmos-12B, DeltaWorld},
      xtick=data,
      xticklabels={Cosmos-4B, Cosmos-12B, \textbf{DeltaWorld-0.3B (Ours)}},
      x tick label style={rotate=22, anchor=north east, font=\sffamily\scriptsize, inner sep=1pt},
      ymajorgrids=true,
      grid style={dotted, gray!60},
      axis line style={black, line width=0.4pt},
      tick style={black},
      legend style={
          at={(0.02, 0.90)},
          anchor=north west,
          legend columns=2,
          draw=gray!40,
          font=\sffamily\scriptsize,
          /tikz/every even column/.append style={column sep=0.3cm}
      }
  ]

  \addlegendimage{line legend, draw=presentline, dotted, line width=2.2pt}
  \addlegendentry{\textit{Present}}

  \addlegendimage{line legend, draw=copylast, dashed, line width=1.8pt}
  \addlegendentry{\textit{Copy last}}

  \draw[copylast, dashed, line width=1.8pt]
      ({rel axis cs:0,0} |- {axis cs:Cosmos-4B, 39.6}) --
      ({rel axis cs:1,0} |- {axis cs:Cosmos-4B, 39.6});

  \draw[presentline, dotted, line width=2.2pt]
      ({rel axis cs:0,0} |- {axis cs:Cosmos-4B, 70.5}) --
      ({rel axis cs:1,0} |- {axis cs:Cosmos-4B, 70.5});

  \addplot[fill=vfmfill, draw=black, line width=0.3pt, area legend] coordinates {
      (Cosmos-4B, 49.1)
      (Cosmos-12B, 53.3)
      (DeltaWorld, 55.4)
  };
  \addlegendentry{\textit{Best}}

  \addplot[fill=tokfill, draw=black, line width=0.3pt, area legend] coordinates {
      (Cosmos-4B, 48.4)
      (Cosmos-12B, 51.2)
      (DeltaWorld, 51.3)
  };
  \addlegendentry{\textit{Mean}}

  \node[font=\sffamily\scriptsize, anchor=south] at (axis cs:Cosmos-4B, 49.5) {60,000 TFLOPs};
  \node[font=\sffamily\scriptsize, anchor=south] at (axis cs:Cosmos-12B, 53.7) {64,000 TFLOPs};
  \node[font=\sffamily\bfseries\scriptsize, anchor=south] at (axis cs:DeltaWorld, 55.8) {31 TFLOPs};

  \end{axis}
  \end{tikzpicture}%
  }%
  \vspace{-2mm}
  \caption{\textbf{Performance comparison.} Compared to the generative world model Cosmos~\cite{agarwal2025cosmos}, our \modelname forecasts futures that better align with real-world outcomes while having over $35\times$ fewer parameters and using $2{,}000\times$ fewer FLOPs.}
  \label{fig:performance_comparison}
\end{figure}

Discriminative world models~\cite{baldassarre2025back,zhou2024dino,karypidis2024dino}, however, produce a single deterministic prediction that, under uncertainty, collapses toward the conditional mean~\cite{walker2016uncertain} rather than capturing distinct future events. Consequently, such models cannot represent the breadth of plausible futures required for reliable downstream decision making. A world model should therefore generate a \emph{set} of plausible future states both accurately and efficiently --- a requirement that naturally calls for a \emph{generative} world model.

Most existing generative world models~\cite{brooks2024sora,agarwal2025cosmos,chen2024diffusion_forcing,hu2023gaia1,bruce2024genie}, however, remain computationally inefficient for three primary reasons:
(i) their representation space is optimized for pixel-level fidelity rather than semantic understanding,
(ii) they require multiple sequential forward passes to produce a single future hypothesis, and
(iii) they fail to exploit the spatio-temporal redundancy that consecutive frames exhibit.
In this work, we take a step toward more efficient generative world modeling by addressing these inefficiencies.

Predicting future world states with pixel-level fidelity is conceptually straightforward but computationally inefficient, as it requires modeling fine-grained visual details that are irrelevant to downstream decision making. For instance, rendering high-fidelity background elements such as trees or buildings provides no actionable information for an autonomous vehicle’s decision to turn left or right. When downstream tasks such as segmentation or depth estimation already operate on certain visual features, prediction can happen directly in that feature space rather than reconstructing human-interpretable pixels. Recent work~\cite{baldassarre2025back, zhou2024dino, karypidis2024dino, walker2025generalist} has consequently shifted toward world models operating in the feature space of vision foundation models (VFMs), demonstrating improved accuracy on downstream dense forecasting tasks while requiring significantly fewer world model parameters than approaches based on pixel reconstruction. However, most of these approaches remain discriminative.

Generative world models can be broadly categorized as discrete~\cite{hu2023gaia1, bruce2024genie} or continuous~\cite{chen2024diffusion_forcing, brooks2024sora, walker2025generalist}. Similar to large language models~\cite{brown2020language}, discrete world models autoregressively predict discrete codes for each spatial position, while continuous world models typically use diffusion denoising over a spatial grid. Both approaches require multiple forward passes per sample, making inference inefficient.

World models typically employ a tokenizer, which encodes frames into a spatio-temporal latent grid that retains a dense correspondence between tokens and frame patches. In natural video, however, consecutive frames differ only in structured and typically low-dimensional ways: backgrounds remain static, and only a small portion of the scene changes between time steps. Representing each frame as a dense spatial feature map results in long context sequences filled with spatially and temporally redundant tokens~\cite{yu2024image,wiegand2003overview}, while requiring the model to predict equally redundant outputs for each future.

Our goal in this work is to develop a generative world model that efficiently generates many diverse futures. To this end, we build on a discriminative world model~\cite{baldassarre2025back} operating in VFM feature space, and make it generative using a simple Best-of-Many (BoM)~\cite{bhattacharyya2018bom} objective: during training, the model generates multiple future hypotheses from different random inputs, and only the one closest to the ground truth is supervised. At inference, this enables the model to map different inputs to different futures in a single forward pass, avoiding iterative denoising~\cite{ho2020denoising}.%

Each sampled future, however, still requires predicting a full spatial feature map under full spatio-temporal context, which is inefficient. We address this with \emph{\tokenizername}, a tokenizer that compresses the change between consecutive frame features into a single continuous \emph{delta token}. By exploiting the low-dimensional structure of temporal change, a single delta token per frame is sufficient to represent consecutive-frame dynamics in VFM feature space, collapsing video from a three-dimensional spatio-temporal representation to a one-dimensional temporal sequence. We combine \tokenizername with the BoM objective to form our world model, \emph{\modelname} (\Cref{fig:teaser}), which operates entirely on these compact sequences of delta tokens. This significantly improves the efficiency of both training and inference. We also observe improved average prediction quality, which we attribute to a natural prior of the delta formulation: predicting no change simply preserves the previous frame, so the model only needs to learn what changes over time.

We evaluate \modelname on unseen evaluation datasets from the dense forecasting benchmark~\cite{baldassarre2025back}, which includes semantic segmentation on VSPW~\cite{miao2021vspw} and Cityscapes~\cite{cordts2016cityscapes}, as well as monocular depth estimation on KITTI~\cite{geiger2013vision}. Following prior work~\cite{baldassarre2025back}, we evaluate both short- and mid-term horizons, using direct prediction for the former and autoregressive rollouts for the latter. Even with its great efficiency, the best predictions from \modelname consistently surpass those of previous generative world models, while producing average predictions competitive with discriminative and generative baselines, confirming that the sampled futures are realistic. Crucially, \modelname achieves this with over $35\times$ fewer parameters and $2{,}000\times$ fewer FLOPs than existing generative world models (\Cref{fig:performance_comparison}), enabling practical downstream applications that rely on efficient generation of diverse futures.

In summary, our contributions are as follows:
\begin{itemize}
\item \textbf{Compressing frame differences to single delta tokens.}
We propose \emph{DeltaTok}, a tokenizer that encodes only the change between consecutive frame features as a single \emph{delta token} (e.g., $1{,}024\times$ fewer tokens at $512\times512$). This removes the need for spatial modeling, reducing video to a purely temporal sequence.
\item \textbf{Efficient generative world modeling.}
We introduce \emph{DeltaWorld}, a compact generative world model that enables efficient generation of multiple plausible futures in a single forward pass, represented as delta tokens.
\end{itemize}

\section{Related Work}
\label{sec:related_work}

\paragraph{Visual tokenization.}
Since the early days of deep learning, images have been compressed and transformed from pixel space into latent space to enable more efficient and effective processing~\cite{hinton2006reducing,vincent2008extracting}. A typical visual tokenizer follows an autoencoder~\cite{hinton2006reducing} architecture and can be broadly categorized into continuous~\cite{kingma2013auto,higgins2017beta,chen2024deep,chen2025softvq,yao2025reconstruction} and discrete~\cite{van2017neural,esser2021taming,mentzer2023finite,yu2023language,yu2024image,weber2024maskbit,kim2025democratizing} designs, depending on whether the latent representation is quantized. These tokenizers are optimized for pixel reconstruction, making them well-suited for visual generation. Alternatively, vision foundation models (VFMs) such as CLIP~\cite{radford2021learning,yu2022coca,zhai2023sigmoid,tschannen2025siglip} or DINO~\cite{caron2021dinov1,oquab2023dinov2,simeoni2025dinov3} can serve as visual tokenizers~\cite{liu2023visual,li2024llava,bai2025qwen2}, providing rich semantic representations better suited for visual understanding, though recent work has shown that such features can also be decoded back to pixels~\cite{zheng2025diffusiontransformersrepresentationautoencoders}.

In this work, we introduce \emph{DeltaTok}, a visual tokenizer that explicitly encodes feature differences between consecutive frames. Unlike existing video tokenization approaches~\cite{li2025adaptok,fan2025reftok}, which are trained to reconstruct pixels, \tokenizername operates in VFM feature space and encodes frame differences into \emph{delta tokens}. While sharing the spirit of classic motion-residual frameworks~\cite{wiegand2003overview} and optical flow~\cite{teed2020raft}, \tokenizername differs fundamentally: it is non-spatial, compressing frame differences into a single semantic token rather than per-pixel motion vectors. This naturally handles occlusions and new objects, where warping-based approaches struggle. Moreover, when temporal redundancy is low, \tokenizername can revert to absolute compression, encoding the new state directly. Together, these properties yield an extremely compact representation that enables efficient generative world modeling.

\paragraph{World modeling.}
Generative modeling for images~\cite{rombach2022high,peebles2023scalable} and videos~\cite{openai_sora_2024,google_veo3_2024} has evolved from early VAE- and GAN-based sampling approaches~\cite{kingma2013auto,goodfellow2014generative} to diffusion~\cite{dhariwal2021diffusion,rombach2022high,peebles2023scalable,lipman2022flow,hoogeboom2023simple,liu2024alleviating,yang20241,he2025flowtok,shin2025deeply} and autoregressive models~\cite{esser2021taming,yu2022scaling,sun2024autoregressive,tian2024visual,yu2025randomized}, as well as hybrid variants integrating multiple paradigms~\cite{li2024autoregressive,ren2024flowar,ren2025beyond,yu2026autoregressive}. These models have achieved remarkable success in producing high-fidelity, aesthetically compelling visual content, demonstrating strong potential for real-world applications~\cite{podell2023sdxl,blattmann2023stable,blackforestlabs_flux_2024,ren2025grouping}.
Beyond high-quality visual synthesis, a growing body of work~\cite{hafner2019dreamer,brooks2024sora,agarwal2025cosmos,chen2024diffusion_forcing,hu2023gaia1,bruce2024genie,micheli2024delta} has focused on constructing \emph{world models} that generate future states of an environment conditioned on past observations and optionally on actions or instructions, aiming to capture the underlying dynamics of the environment.
Approaches operating in VFM feature space~\cite{baldassarre2025back,zhou2024dino,karypidis2024dino,walker2025generalist} (\eg, using DINO features~\cite{oquab2023dinov2}), or learning a predictive feature space end-to-end~\cite{bardes2024vjepa}, further shift world modeling toward semantic structure, reducing the need to model irrelevant pixel-level detail. However, most of these approaches remain non-generative, and thus cannot model diverse futures. More broadly, generative world models, regardless of their representation space, rely on multi-step generation, requiring many forward passes for even a single future. Although some single-pass generative world models exist, they mostly remain task-specific and are not designed for general-purpose forecasting across diverse visual domains~\cite{ha2018worldmodels,hafner2019dreamer,lin2020gswm,emami2020symmetric}.
Building on these insights, we propose \modelname, a compact general-purpose generative world model that represents each frame in VFM feature space as a single token and produces multiple diverse futures in a single forward pass at substantially lower inference cost.

\section{Method}
\label{sec:method}

In this section, we first review the discriminative world model we build on (\Cref{sec:prelim}) and introduce a training objective that extends it into a generative model (\Cref{sec:bom}). We then describe frame-level tokenization (\Cref{sec:frametok}) and subsequently a more targeted variant that compresses only the temporal difference between consecutive frames (\Cref{sec:deltatok}), yielding our proposed \emph{DeltaTok} tokenizer and \emph{DeltaWorld} model.

\subsection{Preliminaries}
\label{sec:prelim}
We build on the discriminative DINO-world~\cite{baldassarre2025back} architecture, which models scene dynamics directly in the feature space of a vision foundation model (VFM). Given VFM features of a set of context frames, the goal is to predict the VFM features of a future frame. Operating in this feature space abstracts away much of the pixel-level variability, allowing a compact predictor to capture temporal dynamics more effectively.

\paragraph{Architecture.}
Given a sequence of $t$ video frames, $V_{1:t} = (v_1, \dots, v_t)$, $v_i \in \mathbb{R}^{H' \times W' \times 3}$, a frozen VFM $\phi$ embeds each frame into a grid of patch tokens: $x_i = \phi(v_i) \in \mathbb{R}^{H \times W \times D}$, where $x_{i,h,w} \in \mathbb{R}^D$ denotes the patch token from frame $i$ at spatial position $(h,w)$. The encoded context is $X_{1:t} = (x_1, \dots, x_t)$, with associated timestamps $T_{1:t} = (\tau_1, \dots, \tau_t)$.
The future predictor $f$ forecasts each patch token $\hat{x}_{t+1,h,w}$ at a target timestamp $\tau_{t+1}$, conditioned on the context. It uses a stack of Transformer blocks~\cite{vaswani2017attention} applying cross-attention from a single learnable query embedding $q$ to the context $X_{1:t}$:
\begin{equation}
\label{eq:predictor}
\hat{x}_{t+1,h,w} = f\!\left(q,\, X_{1:t},\, T_{1:t},\, \tau_{t+1},\, h,\, w\right) \in \mathbb{R}^{D}.
\end{equation}
This operation is performed independently for each spatial location $(h,w)$, with positional embeddings ensuring position-dependent predictions.

\paragraph{Training \& inference.}
Training sequences are constructed by selecting frames at different intervals, using temporal offsets $\Delta\tau$ sampled uniformly from a predefined range, enabling prediction at arbitrary future timestamps. For each sampled timestamp, the nearest video frame is selected and its actual timestamp is used. The predictor is optimized with a smooth L1 loss $\ell$ between predicted and ground-truth features, using teacher forcing~\cite{williams1989learning}: each prediction is conditioned on ground-truth past features $X_{1:t}$, and a causal attention mask restricts it to attending only to earlier frames, enabling all timestamps and context lengths to be predicted in parallel in a single forward pass. At inference, the model can perform an autoregressive rollout, appending $\hat{x}_{t+1}$ to the context before predicting the next.

\subsection{Best-of-Many (BoM) Training}
\label{sec:bom}
DINO-world is a discriminative world model: given a context of previous frame features, it produces a single deterministic prediction of the next frame's features. When the future has multiple plausible outcomes, the regression loss drives the model toward a single averaged prediction~\cite{walker2016uncertain} that may not correspond to any realistic outcome. Hence, it cannot provide the diverse set of plausible futures needed for reliable downstream decision making.

To make the model generative, \ie, capable of sampling multiple plausible futures, we require a mechanism that maps different stochastic inputs to different future hypotheses. Common generative approaches such as diffusion~\cite{ho2020denoising} require multiple forward passes to generate a single sample, which is inefficient. Instead, we adopt a simple \emph{Best-of-Many} (BoM)~\cite{bhattacharyya2018bom} training objective that achieves this in a single forward pass. Concretely, we draw $K$ noise queries from a Gaussian distribution:
\begin{equation}
q^{k} \sim \mathcal{N}(\mu, \Sigma), \qquad k = 1,\dots,K
\end{equation}
each replacing the single learned query $q$ in \eqref{eq:predictor} and shared across all spatial locations $(h,w)$. Using these $K$ queries, the predictor produces $K$ predictions for each spatial location:
\begin{equation}
\hat{x}_{t+1,h,w}^{k}
= f\!\left(q^{k},\, X_{1:t},\, T_{1:t},\, \tau_{t+1},\, h,\, w\right)\in \mathbb{R}^{D}.
\end{equation}
Only the prediction closest to the ground truth is supervised:
\begin{equation}
\begin{split}
&k^\star = \arg\min_{k}
\sum_{h,w}
\ell\!\left(x_{t+1,h,w},\, \hat{x}_{t+1,h,w}^{k}\right); \\
&L_{\text{BoM}}
= \sum_{h,w}
\ell\!\left(x_{t+1,h,w},\, \hat{x}_{t+1,h,w}^{k^\star}\right),
\end{split}
\end{equation}
where $L_{\text{BoM}}$ is the minimized BoM loss. This encourages the model to map different noise queries to different plausible futures directly, preserving the single-pass efficiency of the predictor.

\subsection{Frame Compression to a Single Token}
\label{sec:frametok}
BoM training requires predicting and evaluating many future hypotheses for each context, which becomes expensive when the predictor must output $H \times W$ patch tokens per future under full spatio-temporal context.
To reduce this cost, we compress each frame's feature map into a single \emph{frame token}, reducing the predictor's sequence length from $H \times W$ tokens per frame to one, making the cost of generating many samples negligible. The decoder is then responsible for reconstructing coherent spatial feature maps, simplifying the predictor's task.
Importantly, the BoM loss can now be computed directly in this single-token space, avoiding the need to decode spatial feature maps during predictor training.

\paragraph{Tokenizer architecture.}
We introduce a frame-level tokenizer based on a continuous autoencoder~\cite{hinton2006reducing} design. The encoder $g$ compresses a feature map $x_t$ and a learnable embedding $z_{\mathrm{init}}$ to a single frame token $z_t$:
\begin{equation}
z_t = g(x_t, z_{\mathrm{init}}) \in \mathbb{R}^{D}.
\end{equation}
The tokenizer decoder $h$ reverses this process, reconstructing the feature map from the frame token $z_t$ using $H \times W$ zero-initialized patch tokens $x^{\mathrm{init}}$:
\begin{equation}
\hat{x}_t = h(x^{\mathrm{init}}, z_t).
\end{equation}
Both encoder and decoder are implemented as stacks of Transformer blocks with self-attention.

\paragraph{Tokenizer training.}
The tokenizer is trained separately, before the world model, using a reconstruction loss between the original and reconstructed feature maps:
\begin{equation}
L_{\mathrm{tok}} = \|\,x_t - \hat{x}_t\,\|^2.
\end{equation}
This encourages $z_t$ to serve as a compact representation capturing the information needed to reconstruct $x_t$.

Although frame compression greatly reduces compute, it forces $z_t$ to represent the full scene at each timestep. A single token has limited capacity for faithfully representing each frame's spatial content, and therefore the subtle variations that differentiate one frame from the next, ultimately limiting prediction accuracy.

\begin{figure*}
    \centering
    \resizebox{0.9\linewidth}{!}{\input{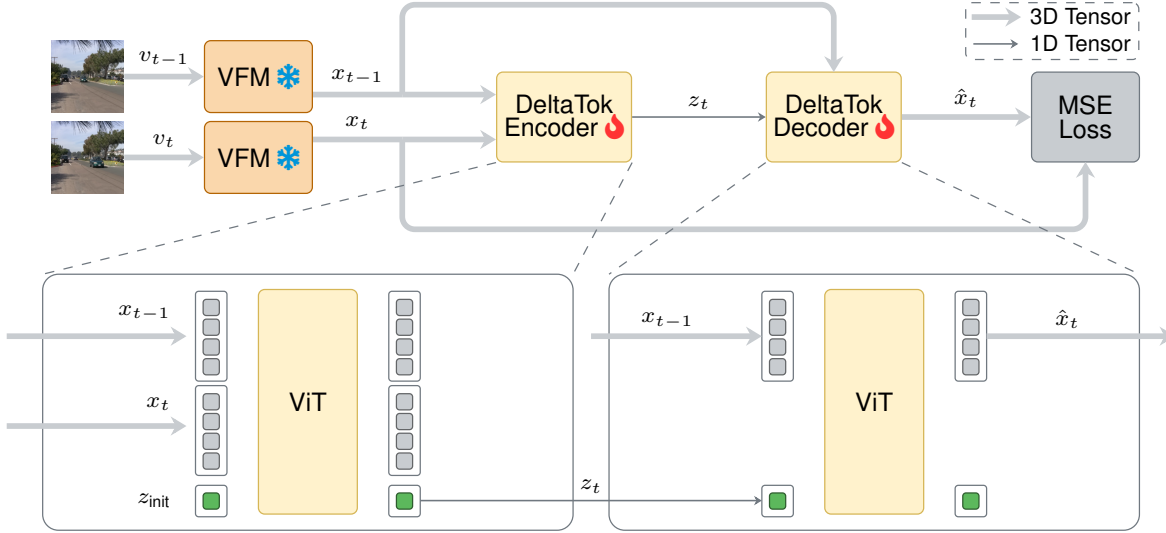}}
    \caption{\textbf{Overview of \tokenizername.} Given two frames encoded by a frozen vision foundation model (VFM) into grids of patch tokens $x_{t-1}$ and $x_t$, the \tokenizername encoder takes both as input and compresses them into a single \emph{delta token} $z_t$. The decoder reconstructs $\hat{x}_{t}$ from $x_{t-1}$ and $z_t$.
    Both encoder and decoder are Vision Transformers (ViT)~\cite{dosovitskiy2021image} trained with a Mean Squared Error (MSE) loss.
    }
    \label{fig:deltatok}
\end{figure*}

\subsection{Delta Compression to a Single Token}
\label{sec:deltatok}

To address the limitations of frame compression, we propose a more targeted approach: compressing only the \emph{change} between consecutive frames in a single token, rather than compressing the entire frame.
The key insight is that $x_t$ differs from $x_{t-1}$ in structured and typically low-dimensional ways, a principle that has long been exploited in video coding through interframe (delta) compression~\cite{wiegand2003overview}. Here we adopt this idea in a different setting: conditioning the tokenizer on the previous frame encourages the single-token representation to encode how to transform the previous frame’s features into the next, which requires significantly less information than re-encoding the entire scene from scratch at each timestep.

\paragraph{DeltaTok.}
We introduce \emph{DeltaTok} (\Cref{fig:deltatok}), which uses the same tokenizer architecture as for frame compression (\Cref{sec:frametok}) but conditions on the previous frame's features. Specifically, the encoder now takes both $x_{t-1}$ and $x_t$ to produce a single \emph{delta token} $z_t$ that encodes the change between them:
\begin{equation}
z_t = g(x_{t-1}, x_t, z_{\mathrm{init}}) \in \mathbb{R}^{D},
\end{equation}
and the decoder now reconstructs the current frame features by \emph{transforming} the previous frame features using the delta token:
\begin{equation}
\hat{x}_t = h(x_{t-1}, z_t).
\end{equation}
\tokenizername is trained using the same reconstruction loss as for frame compression, with frame pairs $(x_{t-1}, x_t)$ drawn from the same uniform timestamp-sampling procedure used for predictor training. As a result, a single delta token can encode changes ranging from near-static scenes, where most of the previous frame can be retained, to large scene transitions, where little can be retained. The inference frame rate controls how much change each token represents.

\begin{figure*}
    \centering
    \resizebox{0.9\linewidth}{!}{\input{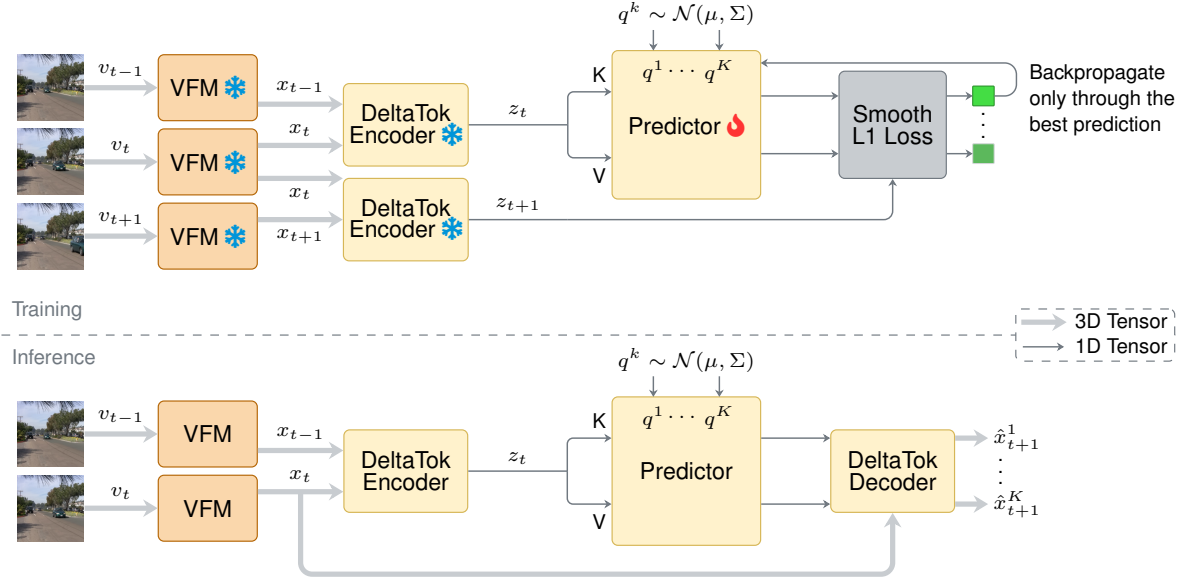}}
    \caption{
    \textbf{Overview of \modelname.}
    The predictor operates entirely on \emph{delta tokens} (Fig.~\ref{fig:deltatok}) rather than spatial tokens, enabling efficient generation of future hypotheses. Best-of-Many training (top) backpropagates only through the best predicted delta token, so that diverse futures can be sampled in a single forward pass at inference (bottom). Shown with two context frames and two queries for illustration.
    }
    \label{fig:deltaworld}
\end{figure*}

\paragraph{DeltaWorld.}
Combining a separately trained, frozen \tokenizername with the future predictor $f$, we obtain \emph{\modelname} (\Cref{fig:deltaworld}), which predicts delta tokens instead of full spatial feature maps. Each input sequence is prepended with a black frame so that the first delta token $z_1$ effectively encodes the absolute features of the first real frame. At each timestep, the predictor operates on the sequence of past delta tokens, $Z_{1:t} = (z_1, \dots, z_t)$, and predicts the next delta token:
\begin{equation}
\hat{z}_{t+1} = f(q^{k}, Z_{1:t}, T_{1:t}, \tau_{t+1}).
\end{equation}
The corresponding spatial feature map can be recovered using the \tokenizername decoder as $\hat{x}_{t+1} = h(x_t, \hat{z}_{t+1})$. During training, each noise query yields a candidate delta token, and the BoM objective selects the best one in delta token space, without requiring decoding. At inference, different noise queries yield diverse future hypotheses in a single forward pass, each representing a plausible evolution of the scene. For autoregressive rollout, the predictor iteratively appends each predicted delta token to the context, operating entirely in delta token space. The decoder can be applied separately to sequentially recover spatial features for downstream tasks.

\tokenizername reduces video from a three-dimensional spatio-temporal representation to a one-dimensional temporal sequence of delta tokens. \modelname operates on this compact sequence, focusing computation on what changes over time and enabling efficient generation of diverse futures.

\section{Experiments}
\label{sec:experiments}

\subsection{Implementation Details}
We perform all experiments in the feature space of the DINOv3~\cite{simeoni2025dinov3} VFM.
We reimplement DINO-world~\cite{baldassarre2025back}, as their code and training data are unavailable.
Following DINO-world, we adopt the ViT-B~\cite{dosovitskiy2021image} variant of the VFM backbone and for simplicity also use the ViT-B configuration for the tokenizer and predictors, though the formulations place no restrictions on scaling.
For the main results in \Cref{tab:results}, both DINO-world and \modelname are trained for 300K iterations with $512\times512$ inputs, and we use $K{=}256$ during BoM training for \modelname. For the ablations in \Cref{tab:steps-results,fig:bom_heatmaps_best_mean}, we use 100K iterations with $256\times256$ inputs, and $K{=}16$ in \Cref{tab:steps-results}. Following DINO-world~\cite{baldassarre2025back}, predictors use a batch size of $1{,}024$, a training sequence length of $8$ frames, and all other predictor training hyperparameters match DINO-world. Predictors are additionally fine-tuned at a 10$\times$ lower learning rate for 5K iterations. The tokenizers are separately trained for 50K iterations at each resolution with a batch size of $1{,}024$. Temporal offsets $\Delta\tau$ are sampled uniformly from $[1/25,\,1/3]$ seconds. Further details are in \Cref{sec:supp:impl_details}.

\subsection{Datasets}

\begin{table}[t]
\centering
\resizebox{\columnwidth}{!}{%
\begin{tabular}{l|ccc}
\toprule
\textbf{Feature} & \textbf{VSPW}~\cite{miao2021vspw} & \textbf{Cityscapes}~\cite{cordts2016cityscapes} & \textbf{KITTI}~\cite{geiger2013vision} \\
\midrule
Task & Segmentation & Segmentation & Depth \\
Domain & Various & Driving & Driving \\
Resolution & Various (HD) & $2{,}048 \times 1{,}024$ & $1{,}216 \times 352$ \\
\# Sequences & 343 & 500 & 28 \\
\bottomrule
\end{tabular}%
}
\caption{\textbf{Evaluation datasets.} We evaluate segmentation and depth at short (${\sim}0.2$\,s) and mid (${\sim}0.6$\,s) prediction horizons.}
\label{tab:dataset_stats}
\end{table}

Similar to the experimental setting of DINO-world~\cite{baldassarre2025back}, we train all models (including our tokenizers) on a large collection of videos spanning diverse domains (${\sim}4\mathrm{M}$ samples; see \Cref{tab:train-dataset-stats}). We also adopt their evaluation datasets (\Cref{tab:dataset_stats}), none of which are included in our training set.

\subsection{Evaluation Settings}
We use the dense forecasting benchmark~\cite{baldassarre2025back}, which evaluates short-term (${\sim}0.2$\,s) and mid-term (${\sim}0.6$\,s) prediction accuracy via segmentation mIoU and depth RMSE on the datasets in \Cref{tab:dataset_stats}. Following the benchmark protocol, a four-frame context is used, with direct prediction for short-term and three-step autoregressive rollout for mid-term evaluation. For our BoM-based models, $K$ futures are rolled out independently, each sampling a fresh query at every step and appending its prediction to its own context. Linear segmentation and depth heads are trained on frozen VFM features, using the training split of each evaluation dataset, following DINO-world~\cite{baldassarre2025back}. These fixed heads are then applied to predicted future spatial feature maps to make segmentation and depth predictions. For the pixel-generating Cosmos baseline~\cite{agarwal2025cosmos}, predicted pixels are re-encoded with the same VFM to ensure feature-level comparability, again matching the protocol of DINO-world~\cite{baldassarre2025back}.

Following~\cite{xu2020vpeg,walker2025generalist}, we draw 20 samples at test time and report both \textit{best} and \textit{mean} scores, unless noted otherwise. The \textit{best} selects the sample closest to the ground truth, reflecting how well the model can produce at least one accurate future within a fixed sample budget. The \textit{mean} averages spatial features across all samples before applying the task head, measuring prediction consistency and enabling fair comparison with discriminative models. For a useful generative world model, both should be strong, as a strong \textit{best} without a strong \textit{mean} may indicate noisy rather than plausible diversity. For measuring FLOPs, we use the DeepSpeed FLOPs Profiler~\cite{rasley2020deepspeed}. Further details are in \Cref{sec:supp:eval_details}.

\subsection{Towards an Efficient Generative World Model}
\label{sec:steps}

As introduced in \Cref{sec:method}, we progressively extend a discriminative world model into an efficient generative one and measure the resulting changes in compute and mid-term forecasting accuracy. In \Cref{sec:supp:eval_details} we provide a detailed FLOPs breakdown of the backbone, tokenizer, and predictor, and in \Cref{sec:supp:disc_abl} we show that delta tokens are also effective in different discriminative world model architectures. %

\paragraph{Step \tablestep{0} – Discriminative baseline.}
We use our reimplementation of the discriminative DINO-world~\cite{baldassarre2025back} architecture as our baseline (\Cref{sec:prelim}), which benefits from operating in VFM feature space rather than a latent space trained for pixel reconstruction. Its performance is reported in \Cref{tab:steps-results}.

\begin{table}[t]
\centering
\small
\renewcommand{\arraystretch}{1.}
\setlength{\tabcolsep}{4pt}
\resizebox{\columnwidth}{!}{%
\begin{tabular}{@{}l r r r l l@{}}
\toprule
\textbf{Step}
& \textbf{GFLOPs $\downarrow$}
& \textbf{Time $\downarrow$}
& \textbf{Mem $\downarrow$}
& \textbf{VSPW $\uparrow$}
& \textbf{Cityscapes $\uparrow$} \\
\midrule
{\color{gray}\tablestep{0} Discriminative}
& {\color{gray}959}
& {\color{gray}$1.0\times$}
& {\color{gray}$1.0\times$}
& {\color{gray}44.8}
& {\color{gray}45.4} \\
\tablestep{1} BoM training
& 12013
& $4.9\times$
& $1.0\times$
& 47.0 (39.4)
& 46.8 (31.1) \\
\tablestep{2} Frame compress
& 6315
& $0.4\times$
& $0.2\times$
& 45.7 (40.3)
& 42.7 (35.5) \\
\tablestep{3} Delta compress
& 6721
& $0.5\times$
& $0.2\times$
& 46.8 (44.4)
& 48.7 (45.5) \\
\bottomrule
\end{tabular}%
}
\caption{\textbf{Towards an efficient generative world model.}
Reporting mid-horizon (${\sim}0.6$\,s) mIoU.
Steps~\tablestep{1-3} use $K{=}16$ during training and report \textit{best}-of-$20$ during evaluation (\textit{mean} in parentheses).
GFLOPs for steps~\tablestep{1-3} reflect generating all 20 samples, and a single prediction for step~\tablestep{0}.
Time and Mem report training time and GPU memory relative to step~\tablestep{0}.
Using $256\times256$ crops.}
\label{tab:steps-results}
\end{table}

\paragraph{Step \tablestep{1} – Best-of-Many (BoM) training.}
To make the baseline generative, we apply the BoM~\cite{bhattacharyya2018bom} objective (\Cref{sec:bom}), conditioning the predictor on noise queries to sample diverse plausible futures. As shown in \Cref{tab:steps-results}, this enables the model to produce at least one noticeably more accurate future within a fixed budget of 20 samples. However, the \textit{mean} prediction drops sharply (from 45.4 to 31.1 on Cityscapes and from 44.8 to 39.4 on VSPW). We observe many samples collapsing to degenerate predictions, \eg, a single semantic class for the entire frame. In addition, predicting multiple futures increases training time by roughly $5\times$, even when using only $K{=}16$ during training. At inference, the predictor accounts for 97\% of total FLOPs when generating 20 samples, as it must predict full spatial feature maps for each (\Cref{tab:supp:efficiency}).

\begin{table*}
\small
\centering
\renewcommand{\arraystretch}{1.2}
\setlength{\tabcolsep}{4pt}
\begin{tabularx}{\textwidth}{l c XXXXXX}
\toprule
& \textbf{GFLOPs $\downarrow$} 
& \multicolumn{2}{c}{\textbf{VSPW mIoU $\uparrow$}} 
& \multicolumn{2}{c}{\textbf{Cityscapes mIoU $\uparrow$}} 
& \multicolumn{2}{c}{\textbf{KITTI RMSE $\downarrow$}} \\
\cmidrule(lr){3-4} \cmidrule(lr){5-6} \cmidrule(lr){7-8}
&  & Short & Mid & Short & Mid & Short & Mid \\
\midrule
\textit{Copy last (lower bound)}
& 
& \textit{51.2} & \textit{44.3}
& \textit{53.5} & \textit{39.6}
& \textit{3.76} & \textit{4.86} \\
\midrule
\textcolor{gray}{DINO-world$^\dagger$~\cite{baldassarre2025back}}
& \textcolor{gray}{$5.8{\times}10^{3}$}
& \textcolor{gray}{54.0} & \textcolor{gray}{47.9}
& \textcolor{gray}{62.0} & \textcolor{gray}{49.8}
& \textcolor{gray}{3.16} & \textcolor{gray}{4.07} \\
Cosmos-4B$^\ddagger$~\cite{agarwal2025cosmos}
& $6.0{\times}10^{7}$
& 51.1 (49.7) & 47.0 (44.5)
& 55.1 (54.9) & 49.1 (48.4)
& 3.82 (3.75) & 4.08 (4.14) \\
Cosmos-12B$^\ddagger$~\cite{agarwal2025cosmos}
& $6.4{\times}10^{7}$
& 51.7 (50.7) & 47.7 (45.5)
& 55.3 (56.0) & 53.3 (51.2)
& 3.72 (3.71) & 4.01 (4.14) \\
\textbf{DeltaWorld (Ours)} 
& $3.1{\times}10^{4}$
& \textbf{55.4} (53.7) & \textbf{50.1} (46.7)
& \textbf{65.8} (63.9) & \textbf{55.4} (51.3)
& \textbf{3.00} (3.17) & \textbf{3.88} (4.17) \\
\midrule
\textit{Present (upper bound)}
&
& \textit{58.4} & \textit{58.4}
& \textit{70.5} & \textit{70.5}
& \textit{2.79} & \textit{2.79} \\
\bottomrule
\end{tabularx}
\caption{\textbf{Dense forecasting.}
Reporting short (${\sim}0.2$\,s, direct) and mid (${\sim}0.6$\,s, 3-step rollout) prediction horizons. Generative models report \textit{best}-of-$20$ evaluation (\textit{mean} in parentheses). GFLOPs reflect generating all 20 samples for generative models and a single prediction for DINO-world.
Using $512\times512$ crops.
$^\dagger$Our reimplementation.
$^\ddagger$Both variants use another 7B diffusion decoder, dominating FLOPs.}
\label{tab:results}
\end{table*}

\paragraph{Step \tablestep{2} – Frame compression.}
To improve efficiency and simplify prediction, we train a tokenizer (\Cref{sec:frametok}) that compresses each frame’s spatial feature map (256 tokens at $256\times256$ inputs) into a single frame token, and perform world modeling directly in this compressed space. \Cref{tab:steps-results} shows that frame compression makes BoM sampling more than an order of magnitude faster than in step~\tablestep{1}, even outpacing the discriminative baseline, while also using $5\times$ less memory. This is because both the context and predictions are now single tokens, and the BoM loss is computed directly in frame token space rather than in the full spatial feature space. In terms of accuracy, the \textit{mean} prediction improves over step~\tablestep{1}. We hypothesize that the tokenizer decoder, trained to reconstruct coherent feature maps, makes it harder for samples to collapse to degenerate predictions, though accuracy remains well below the discriminative baseline. Representing an entire frame with a single token limits the representational capacity, which may ultimately lower both \textit{best} and \textit{mean} predictions.

\paragraph{Step \tablestep{3} – Delta compression (DeltaWorld).}
To address the limitations of full frame compression, we encode only the change between consecutive frames as a single \emph{delta token} using \tokenizername (\Cref{sec:deltatok}). Because the delta captures only the information needed to transform $x_{t-1}$ into $x_t$, it can be represented more accurately in a single token. This yields our final model, \modelname, which predicts delta tokens rather than full spatial features or frame tokens.
As shown in \Cref{tab:steps-results}, \modelname substantially improves over step~\tablestep{2} on both \textit{best} and \textit{mean} metrics, confirming the benefit of compressing only temporal differences rather than full frames. As in step~\tablestep{2}, \modelname operates on only a single token per frame, so the predictor accounts for just 0.5\% of total inference FLOPs when generating 20 samples (\Cref{tab:supp:efficiency}). Additionally, \modelname's \textit{best} predictions match or exceed BoM without any compression in step~\tablestep{1} (+1.9\,mIoU on Cityscapes, within 0.2\,mIoU on VSPW), while its \textit{mean} mIoU recovers to the level of the discriminative baseline optimized for mean prediction in step~\tablestep{0} (44.4 \vs 44.8 on VSPW and 45.5 \vs 45.4 on Cityscapes). We attribute the recovered \textit{mean} to a natural prior of the delta formulation: predicting no change simply preserves the previous frame. These results demonstrate that combining BoM training with delta compression achieves our goal of an efficient generative world model that produces diverse, plausible futures.

\subsection{Best-of-Many Sample Scaling}

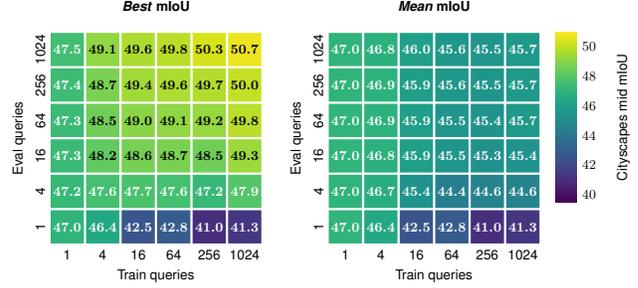
\begin{figure}
  \centering
  \resizebox{\linewidth}{!}{%
  \begin{tikzpicture}
  \pgfplotsset{
      heatmap axis/.style={
          width=5cm, height=5cm,
          scale only axis,
          xmin=-0.5, xmax=5.5,
          ymin=-0.5, ymax=5.5,
          xtick={0,1,2,3,4,5},
          xticklabels={1, 4, 16, 64, 256, 1024},
          ytick={0,1,2,3,4,5},
          yticklabels={1024, 256, 64, 16, 4, 1}, 
          xlabel={Train queries},
          ylabel={Eval queries},
          title style={font=\sffamily\bfseries\small, yshift=1ex},
          xlabel style={font=\sffamily\small},
          ylabel style={font=\sffamily\small},
          tick label style={font=\sffamily\small},
          yticklabel style={rotate=90, anchor=center, yshift=6pt},
          axis line style={draw=none},
          tick style={draw=none},
          enlargelimits=false,
          axis on top,
          colormap name=viridis,
          point meta min=39.5,
          point meta max=51.0,
          every node near coord/.append style={
              font=\sffamily\boldmath\small,
              anchor=center,
              /pgf/number format/fixed,
              /pgf/number format/fixed zerofill,
              /pgf/number format/precision=1
          },
          nodes near coords={%
              \pgfmathfloattofixed{\pgfplotspointmeta}%
              \pgfmathparse{\pgfmathresult < 48 ? 1 : 0}%
              \ifnum\pgfmathresult=1\relax
                  \color{white}%
              \else
                  \color{black}%
              \fi
              \pgfmathprintnumber{\pgfplotspointmeta}%
          }
      }
  }

  \begin{axis}[
      name=ax1,
      heatmap axis,
      title={\textit{Best} mIoU},
  ]

  \addplot [
      matrix plot,
      mesh/cols=6,
      point meta=explicit
  ] coordinates {
      (0,0) [47.5] (1,0) [49.1] (2,0) [49.6] (3,0) [49.8] (4,0) [50.3] (5,0) [50.7]
      (0,1) [47.4] (1,1) [48.7] (2,1) [49.4] (3,1) [49.6] (4,1) [49.7] (5,1) [50.0]
      (0,2) [47.3] (1,2) [48.5] (2,2) [49.0] (3,2) [49.1] (4,2) [49.2] (5,2) [49.8]
      (0,3) [47.3] (1,3) [48.2] (2,3) [48.6] (3,3) [48.7] (4,3) [48.5] (5,3) [49.3]
      (0,4) [47.2] (1,4) [47.6] (2,4) [47.7] (3,4) [47.6] (4,4) [47.2] (5,4) [47.9]
      (0,5) [47.0] (1,5) [46.4] (2,5) [42.5] (3,5) [42.8] (4,5) [41.0] (5,5) [41.3]
  };

  \draw[white, line width=1.5pt] (-0.5, 0.5) -- (5.5, 0.5);
  \draw[white, line width=1.5pt] (-0.5, 1.5) -- (5.5, 1.5);
  \draw[white, line width=1.5pt] (-0.5, 2.5) -- (5.5, 2.5);
  \draw[white, line width=1.5pt] (-0.5, 3.5) -- (5.5, 3.5);
  \draw[white, line width=1.5pt] (-0.5, 4.5) -- (5.5, 4.5);
  \draw[white, line width=1.5pt] (0.5, -0.5) -- (0.5, 5.5);
  \draw[white, line width=1.5pt] (1.5, -0.5) -- (1.5, 5.5);
  \draw[white, line width=1.5pt] (2.5, -0.5) -- (2.5, 5.5);
  \draw[white, line width=1.5pt] (3.5, -0.5) -- (3.5, 5.5);
  \draw[white, line width=1.5pt] (4.5, -0.5) -- (4.5, 5.5);
  \draw[white, line width=1.5pt] (-0.5,-0.5) rectangle (5.5,5.5);
  \end{axis}

  \begin{axis}[
      at={(ax1.outer east)}, anchor=outer west,
      xshift=0.5cm,
      name=ax2,
      heatmap axis,
      title={\textit{Mean} mIoU},
      colorbar,
      colorbar style={
          height=4cm,
          ylabel={Cityscapes mid mIoU},
          ylabel style={font=\sffamily\small, yshift=-0.25cm},
          tick label style={font=\sffamily\small},
          ytick={40, 42, 44, 46, 48, 50},
          axis line style={draw=none},
          tick style={draw=none}
      }
  ]

  \addplot [
      matrix plot,
      mesh/cols=6,
      point meta=explicit
  ] coordinates {
      (0,0) [47.0] (1,0) [46.8] (2,0) [46.0] (3,0) [45.6] (4,0) [45.5] (5,0) [45.7]
      (0,1) [47.0] (1,1) [46.9] (2,1) [45.9] (3,1) [45.6] (4,1) [45.5] (5,1) [45.7]
      (0,2) [47.0] (1,2) [46.9] (2,2) [45.9] (3,2) [45.5] (4,2) [45.4] (5,2) [45.7]
      (0,3) [47.0] (1,3) [46.8] (2,3) [45.9] (3,3) [45.5] (4,3) [45.3] (5,3) [45.4]
      (0,4) [47.0] (1,4) [46.7] (2,4) [45.4] (3,4) [44.4] (4,4) [44.6] (5,4) [44.6]
      (0,5) [47.0] (1,5) [46.4] (2,5) [42.5] (3,5) [42.8] (4,5) [41.0] (5,5) [41.3]
  };

  \draw[white, line width=1.5pt] (-0.5, 0.5) -- (5.5, 0.5);
  \draw[white, line width=1.5pt] (-0.5, 1.5) -- (5.5, 1.5);
  \draw[white, line width=1.5pt] (-0.5, 2.5) -- (5.5, 2.5);
  \draw[white, line width=1.5pt] (-0.5, 3.5) -- (5.5, 3.5);
  \draw[white, line width=1.5pt] (-0.5, 4.5) -- (5.5, 4.5);
  \draw[white, line width=1.5pt] (0.5, -0.5) -- (0.5, 5.5);
  \draw[white, line width=1.5pt] (1.5, -0.5) -- (1.5, 5.5);
  \draw[white, line width=1.5pt] (2.5, -0.5) -- (2.5, 5.5);
  \draw[white, line width=1.5pt] (3.5, -0.5) -- (3.5, 5.5);
  \draw[white, line width=1.5pt] (4.5, -0.5) -- (4.5, 5.5);
  \draw[white, line width=1.5pt] (-0.5,-0.5) rectangle (5.5,5.5);
  \end{axis}

  \end{tikzpicture}%
  }
  \caption{\textbf{Best-of-Many sample scaling.} Effect of the number of training and evaluation queries on Cityscapes mid-horizon (${\sim}0.6$\,s) mIoU.
  Using $256\times256$ crops.
  }
  \label{fig:bom_heatmaps_best_mean}
\end{figure}

The Best-of-Many (BoM) objective introduces a hyperparameter $K$ that controls how many queries are sampled during training. \Cref{fig:bom_heatmaps_best_mean} shows how increasing $K$ affects the \textit{best} and \textit{mean} scores for different numbers of evaluation queries. The \textit{best} score generally improves for any fixed number of evaluation queries ($>1$), with no sign of saturation. This indicates that the model keeps learning to predict more specific and accurate futures as $K$ grows. Increasing $K$ modestly lowers the \textit{mean} score but stabilizes beyond $K{=}64$ (with $>1$ evaluation queries), indicating more diversity does not come at the cost of average prediction quality. Together with delta compression (\Cref{sec:steps}), these results show that BoM provides a simple but effective way to extend a discriminative world model into an efficient generative one.

\subsection{Dense Forecasting Benchmark}

We compare our model to prior world models on the dense forecasting benchmark~\cite{baldassarre2025back}. Since no public general-purpose generative world models operate in VFM feature space, we follow the benchmark’s generative baselines: two sizes of Cosmos~\cite{agarwal2025cosmos}, a generative world model operating in a latent space trained for pixel reconstruction. We also report the discriminative DINO-world~\cite{baldassarre2025back}, trained on the same data as \modelname. As lower and upper bounds, \textit{Copy last} repeats the last observed frame’s features as the prediction, while \textit{Present} uses the ground-truth future frame’s features.

Results in \Cref{tab:results} show that despite Cosmos using roughly $2{,}000\times$ more FLOPs, its performance generally lags behind \modelname, with \modelname’s \textit{best} surpassing that of Cosmos across all metrics, while achieving stronger \textit{mean} scores across nearly all metrics. This suggests that modeling temporal differences in a frozen VFM’s feature space allows a significantly simpler generative model to align more closely with real future modes, while generalizing to diverse domains such as VSPW. This also demonstrates that producing diverse samples does not necessarily require multiple forward passes. In fact, the gap between \modelname’s \textit{best} and \textit{mean} scores is consistently larger than that of Cosmos, indicating more meaningful sample diversity.

Compared to the single prediction of the discriminative DINO-World, \modelname's \textit{mean} scores are modestly better on Cityscapes and modestly worse on VSPW and KITTI. As expected, the best of its multiple samples substantially outperforms the single deterministic prediction. Together, this shows the sampled futures cover realistic modes a deterministic model cannot capture.
\begin{figure}
    \centering
    \includegraphics[width=\linewidth]{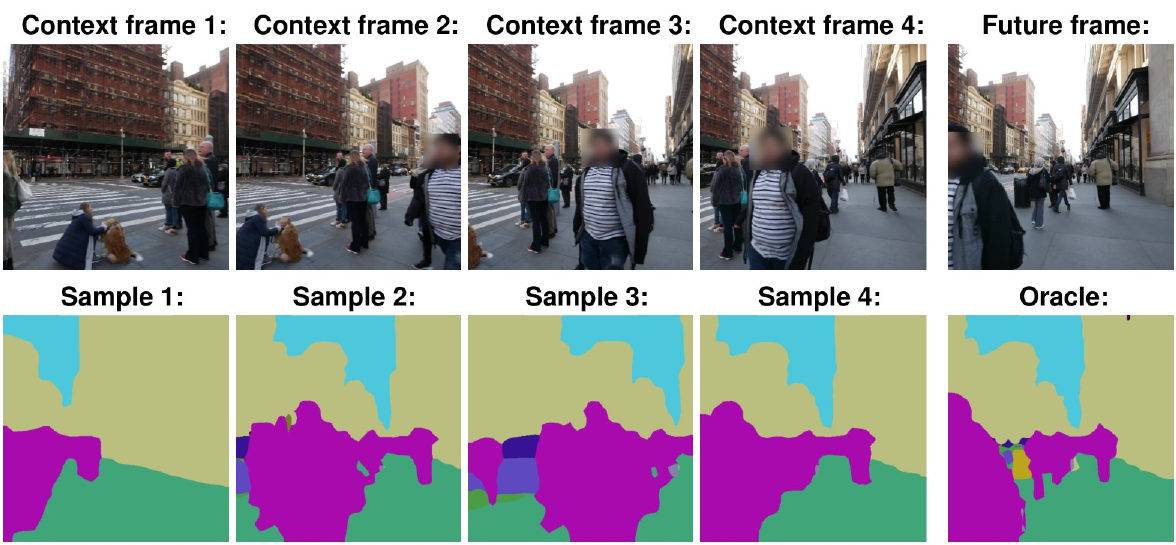}
    \caption{\textbf{Diverse sampled futures.} Top row: four context frames and the future frame. Bottom row: four sampled \modelname predictions and the oracle. In this VSPW~\cite{miao2021vspw} example, the pedestrian's position and ego-camera motion lead to multiple plausible futures.}%
    \label{fig:diverse_samples}
\end{figure}
\Cref{fig:diverse_samples} visualizes this diversity: given four context frames, \modelname produces futures that differ in the pedestrian's position and ego-camera motion. We provide additional qualitative examples in \Cref{sec:supp:qualitative}.

These results demonstrate that representing video with \emph{delta tokens} enables an efficient generative world model that is competitive with the discriminative baseline while outperforming generative models across nearly all metrics.

\section{Conclusion}
\label{sec:conclusion}

In this work, we present \tokenizername, a video tokenizer that encodes the change between consecutive frames as a single delta token, and introduce \modelname, an efficient generative world model built on this representation. \modelname generates multiple diverse yet plausible futures in a single forward pass at orders-of-magnitude lower compute than prior generative world models.
By replacing costly spatial feature maps with delta tokens, \modelname focuses solely on temporal change, boosting both speed and accuracy. This lays the groundwork for scaling predictor size, context length, and rollout depth.\footnote{We discuss limitations and future directions in \Cref{sec:supp:limitations}.} By demonstrating that videos can be represented using only the temporal dimension, delta tokens offer a compact representation for video understanding and generation at scale.

\clearpage
{
    \small
    \bibliographystyle{ieeenat_fullname}
    \bibliography{main}

@String(CVPR= {IEEE Conf. Comput. Vis. Pattern Recog.})

@String(ICCV= {Int. Conf. Comput. Vis.})

@String(ECCV= {Eur. Conf. Comput. Vis.})

@String(NIPS= {Adv. Neural Inform. Process. Syst.})

@String(ICLR = {Int. Conf. Learn. Represent.})

@String(CVPR  = {CVPR})

@String(ICCV  = {ICCV})

@String(ECCV  = {ECCV})

@String(NIPS  = {NeurIPS})

@String(ICLR  = {ICLR})

@String(TMLR  = {TMLR})

@article{agarwal2025cosmos,
  title={{Cosmos World Foundation Model Platform for Physical AI}},
  author={Agarwal, Niket and Ali, Arslan and Bala, Maciej and Balaji, Yogesh and Barker, Erik and Cai, Tiffany and Chattopadhyay, Prithvijit and Chen, Yongxin and Cui, Yin and Ding, Yifan and others},
  journal={arXiv preprint arXiv:2501.03575},
  year={2025}
}

@article{bai2025qwen2,
  title={{Qwen2.5-VL Technical Report}},
  author={Bai, Shuai and Chen, Keqin and Liu, Xuejing and Wang, Jialin and Ge, Wenbin and Song, Sibo and Dang, Kai and Wang, Peng and Wang, Shijie and Tang, Jun and others},
  journal={arXiv preprint arXiv:2502.13923},
  year={2025}
}

@inproceedings{baldassarre2025back,
  title={{Back to the Features: DINO as a Foundation for Video World Models}},
  author={Baldassarre, Federico and Szafraniec, Marc and Terver, Basile and Khalidov, Vasil and Massa, Francisco and LeCun, Yann and Labatut, Patrick and Seitzer, Maximilian and Bojanowski, Piotr},
  booktitle={ICML},
  year={2025}
}

@article{bardes2024vjepa,
  title={{Revisiting Feature Prediction for Learning Visual Representations from Video}},
  author={Bardes, Adrien and Garrido, Quentin and Ponce, Jean and Chen, Xinlei and Rabbat, Michael and LeCun, Yann and Assran, Mido and Ballas, Nicolas},
  journal=TMLR,
  year={2024}
}

@inproceedings{bhattacharyya2018bom,
  title     = {{Accurate and Diverse Sampling of Sequences Based on a ``Best of Many'' Sample Objective}},
  author    = {Bhattacharyya, Apratim and Schiele, Bernt and Fritz, Mario},
  booktitle = CVPR,
  year      = {2018}
}

@misc{blackforestlabs_flux_2024,
  author       = {{Black Forest Labs}},
  title        = {{FLUX}},
  year         = {2024},
  howpublished = {\url{https://github.com/black-forest-labs/flux}}
}

@article{blattmann2023stable,
  title={{Stable Video Diffusion: Scaling Latent Video Diffusion Models to Large Datasets}},
  author={Blattmann, Andreas and Dockhorn, Tim and Kulal, Sumith and Mendelevitch, Daniel and Kilian, Maciej and Lorenz, Dominik and Levi, Yam and English, Zion and Voleti, Vikram and Letts, Adam and others},
  journal={arXiv preprint arXiv:2311.15127},
  year={2023}
}

@article{brooks2024sora,
  title={{Video Generation Models as World Simulators}},
  author={Brooks, Tim and Peebles, Bill and Holmes, Connor and DePue, Will and Guo, Yufei and Jing, Li and Schnurr, David and Taylor, Joe and Luhman, Troy and Luhman, Eric and others},
  journal={OpenAI Blog},
  volume={1},
  number={8},
  pages={1},
  year={2024}
}

@article{brown2020language,
  title={{Language Models are Few-Shot Learners}},
  author={Brown, Tom and Mann, Benjamin and Ryder, Nick and Subbiah, Melanie and Kaplan, Jared D and Dhariwal, Prafulla and Neelakantan, Arvind and Shyam, Pranav and Sastry, Girish and Askell, Amanda and others},
  journal=NIPS,
  year={2020}
}

@inproceedings{bruce2024genie,
  title={{Genie: Generative Interactive Environments}},
  author={Bruce, Jake and Dennis, Michael D and Edwards, Ashley and Parker-Holder, Jack and Shi, Yuge and Hughes, Edward and Lai, Matthew and Mavalankar, Aditi and Steigerwald, Richie and Apps, Chris and others},
  booktitle={ICML},
  year={2024}
}

@inproceedings{caron2021dinov1,
  title={{Emerging Properties in Self-Supervised Vision Transformers}},
  author={Caron, Mathilde and Touvron, Hugo and Misra, Ishan and J{\'e}gou, Herv{\'e} and Mairal, Julien and Bojanowski, Piotr and Joulin, Armand},
  booktitle={ICCV},
  year={2021}
}

@inproceedings{chen2024deep,
  title={{Deep Compression Autoencoder for Efficient High-Resolution Diffusion Models}},
  author={Chen, Junyu and Cai, Han and Chen, Junsong and Xie, Enze and Yang, Shang and Tang, Haotian and Li, Muyang and Lu, Yao and Han, Song},
  booktitle={ICLR},
  year={2025}
}

@article{chen2024diffusion_forcing,
  title        = {{Diffusion Forcing: Next-token Prediction Meets Full-Sequence Diffusion}},
  author       = {Chen, Boyuan and Mart{\'\i} Mons{\'o}, Diego and Du, Yilun and Simchowitz, Max and Tedrake, Russ and Sitzmann, Vincent},
  journal    = {NeurIPS},
  year         = {2024}
}

@inproceedings{chen2025softvq,
  title={{SoftVQ-VAE: Efficient 1-Dimensional Continuous Tokenizer}},
  author={Chen, Hao and Wang, Ze and Li, Xiang and Sun, Ximeng and Chen, Fangyi and Liu, Jiang and Wang, Jindong and Raj, Bhiksha and Liu, Zicheng and Barsoum, Emad},
  booktitle={CVPR},
  year={2025}
}

@inproceedings{cordts2016cityscapes,
  title={{The Cityscapes Dataset for Semantic Urban Scene Understanding}},
  author={Cordts, Marius and Omran, Mohamed and Ramos, Sebastian and Rehfeld, Timo and Enzweiler, Markus and Benenson, Rodrigo and Franke, Uwe and Roth, Stefan and Schiele, Bernt},
  booktitle={CVPR},
  year={2016}
}

@inproceedings{deng2009imagenet,
  title={{ImageNet: A Large-Scale Hierarchical Image Database}},
  author={Deng, Jia and Dong, Wei and Socher, Richard and Li, Li-Jia and Li, Kai and Fei-Fei, Li},
  booktitle=CVPR,
  year={2009}
}

@article{dhariwal2021diffusion,
  title={{Diffusion Models Beat GANs on Image Synthesis}},
  author={Dhariwal, Prafulla and Nichol, Alexander},
  journal=NIPS,
  year={2021}
}

@inproceedings{dosovitskiy2021image,
  title={{An Image is Worth 16x16 Words: Transformers for Image Recognition at Scale}},
  author={Dosovitskiy, Alexey},
  booktitle={ICLR},
  year={2021}
}

@inproceedings{eigen2014depth,
  title={Depth map prediction from a single image using a multi-scale deep network},
  author={Eigen, David and Puhrsch, Christian and Fergus, Rob},
  booktitle={NeurIPS},
  year={2014}
}

@inproceedings{emami2020symmetric,
  title     = {{A Symmetric and Object-Centric World Model for Stochastic Environments}},
  author    = {Emami, Patrick and He, Pan and Rangarajan, Anand and Ranka, Sanjay},
  booktitle = {NeurIPS Workshop on Object Representations for Learning and Reasoning},
  year      = {2020}
}

@inproceedings{esser2021taming,
  title={{Taming Transformers for High-Resolution Image Synthesis}},
  author={Esser, Patrick and Rombach, Robin and Ommer, Bjorn},
  booktitle={CVPR},
  year={2021}
}

@article{fan2025reftok,
  title={{RefTok: Reference-Based Tokenization for Video Generation}},
  author={Fan, Xiang and Sun, Xiaohang and Thakkar, Kushan and Liu, Zhu and Bhat, Vimal and Krishna, Ranjay and Hao, Xiang},
  journal={arXiv preprint arXiv:2507.02862},
  year={2025}
}

@inproceedings{garg2016unsupervised,
  title={Unsupervised {CNN} for single view depth estimation: Geometry to the rescue},
  author={Garg, Ravi and Bg, Vijay Kumar and Carneiro, Gustavo and Reid, Ian},
  booktitle={ECCV},
  year={2016}
}

@article{geiger2013vision,
  title={{Vision meets Robotics: The KITTI Dataset}},
  author={Geiger, Andreas and Lenz, Philip and Stiller, Christoph and Urtasun, Raquel},
  journal={The international journal of robotics research},
  volume={32},
  number={11},
  pages={1231--1237},
  year={2013}
}

@article{goodfellow2014generative,
  title={{Generative Adversarial Nets}},
  author={Goodfellow, Ian J and Pouget-Abadie, Jean and Mirza, Mehdi and Xu, Bing and Warde-Farley, David and Ozair, Sherjil and Courville, Aaron and Bengio, Yoshua},
  journal=NIPS,
  year={2014}
}

@misc{google_veo3_2024,
  title        = {{Veo 3}},
  author       = {{Google DeepMind}},
  year         = {2025},
  howpublished = {\url{https://deepmind.google/veo}}
}

@article{ha2018worldmodels,
  title     = {{Recurrent World Models Facilitate Policy Evolution}},
  author    = {Ha, David and Schmidhuber, J{\"u}rgen},
  journal = NIPS,
  year      = {2018}
}

@inproceedings{hafner2019dreamer,
  title={{Dream to Control: Learning Behaviors by Latent Imagination}},
  author={Hafner, Danijar and Lillicrap, Timothy and Ba, Jimmy and Norouzi, Mohammad},
  booktitle=ICLR,
  year={2020}
}

@inproceedings{he2025flowtok,
  title={{FlowTok: Flowing Seamlessly Across Text and Image Tokens}},
  author={He, Ju and Yu, Qihang and Liu, Qihao and Chen, Liang-Chieh},
  booktitle={ICCV},
  year={2025}
}

@inproceedings{higgins2017beta,
  title={{$\beta$-VAE: Learning Basic Visual Concepts with a Constrained Variational Framework}},
  author={Higgins, Irina and Matthey, Loic and Pal, Arka and Burgess, Christopher and Glorot, Xavier and Botvinick, Matthew and Mohamed, Shakir and Lerchner, Alexander},
  booktitle={ICLR},
  year={2017}
}

@article{hinton2006reducing,
  title={{Reducing the Dimensionality of Data with Neural Networks}},
  author={Hinton, Geoffrey E and Salakhutdinov, Ruslan R},
  journal={science},
  volume={313},
  number={5786},
  pages={504--507},
  year={2006},
  publisher={American Association for the Advancement of Science}
}

@article{ho2020denoising,
  title={{Denoising Diffusion Probabilistic Models}},
  author={Ho, Jonathan and Jain, Ajay and Abbeel, Pieter},
  journal={NeurIPS},
  year={2020}
}

@inproceedings{hoogeboom2023simple,
  title={{simple diffusion: End-to-end diffusion for high resolution images}},
  author={Hoogeboom, Emiel and Heek, Jonathan and Salimans, Tim},
  booktitle={ICML},
  year={2023}
}

@article{hu2023gaia1,
  title        = {{GAIA-1: A Generative World Model for Autonomous Driving}},
  author       = {Hu, Anthony and Russell, Lloyd and Yeo, Hudson and Murez, Zak and Fedoseev, George and Kendall, Alex and Shotton, Jamie and Corrado, Gianluca},
  journal      = {arXiv preprint arXiv:2309.17080},
  year         = {2023}
}

@inproceedings{karypidis2024dino,
  title={{DINO-Foresight: Looking into the Future with DINO}},
  author={Efstathios Karypidis and Ioannis Kakogeorgiou and Spyros Gidaris and Nikos Komodakis},
  booktitle={NeurIPS},
  year={2025}
}

@inproceedings{kim2025democratizing,
  title={{Democratizing Text-to-Image Masked Generative Models with Compact Text-Aware One-Dimensional Tokens}},
  author={Kim, Dongwon and He, Ju and Yu, Qihang and Yang, Chenglin and Shen, Xiaohui and Kwak, Suha and Chen, Liang-Chieh},
  booktitle={ICCV},
  year={2025}
}

@inproceedings{kingma2013auto,
  title={{Auto-Encoding Variational Bayes}},
  author={Kingma, Diederik P and Welling, Max},
  booktitle={ICLR},
  year={2014}
}

@article{li2024autoregressive,
  title={{Autoregressive Image Generation Without Vector Quantization}},
  author={Li, Tianhong and Tian, Yonglong and Li, He and Deng, Mingyang and He, Kaiming},
  journal=NIPS,
  year={2024}
}

@article{li2024llava,
  title={{LLaVA-OneVision: Easy Visual Task Transfer}},
  author={Li, Bo and Zhang, Yuanhan and Guo, Dong and Zhang, Renrui and Li, Feng and Zhang, Hao and Zhang, Kaichen and Zhang, Peiyuan and Li, Yanwei and Liu, Ziwei and others},
  journal=TMLR,
  year={2025}
}

@article{li2025adaptok,
  title   = {{Learning Adaptive and Temporally Causal Video Tokenization in a 1D Latent Space}},
  author  = {Li, Yan and Tian, Changyao and Xia, Renqiu and Liao, Ning and Guo, Weiwei and Yan, Junchi and Li, Hongsheng and Dai, Jifeng and Li, Hao and Yang, Xue},
  journal = {arXiv preprint arXiv:2505.17011},
  year    = {2025}
}

@inproceedings{lin2020gswm,
  title     = {{Improving Generative Imagination in Object-Centric World Models}},
  author    = {Lin, Zhixuan and Wu, Yi-Fu and Peri, Skand Vishwanath and Fu, Bofeng and Jiang, Jindong and Ahn, Sungjin},
  booktitle = {ICML},
  year      = {2020}
}

@inproceedings{lipman2022flow,
  title={{Flow Matching for Generative Modeling}},
  author={Lipman, Yaron and Chen, Ricky TQ and Ben-Hamu, Heli and Nickel, Maximilian and Le, Matt},
  booktitle={ICLR},
  year={2023}
}

@article{liu2023visual,
  title={{Visual Instruction Tuning}},
  author={Liu, Haotian and Li, Chunyuan and Wu, Qingyang and Lee, Yong Jae},
  journal=NIPS,
  year={2023}
}

@article{liu2024alleviating,
  title={{Alleviating Distortion in Image Generation via Multi-Resolution Diffusion Models and Time-Dependent Layer Normalization}},
  author={Liu, Qihao and Zeng, Zhanpeng and He, Ju and Yu, Qihang and Shen, Xiaohui and Chen, Liang-Chieh},
  journal=NIPS,
  year={2024}
}

@inproceedings{loshchilov2019adamw,
  title        = {{Decoupled Weight Decay Regularization}},
  author       = {Ilya Loshchilov and Frank Hutter},
  year         = 2019,
  booktitle    = ICLR
}

@inproceedings{mentzer2023finite,
  title={{Finite Scalar Quantization: VQ-VAE Made Simple}},
  author={Mentzer, Fabian and Minnen, David and Agustsson, Eirikur and Tschannen, Michael},
  booktitle={ICLR},
  year={2024}
}

@inproceedings{miao2021vspw,
  title={{VSPW: A Large-scale Dataset for Video Scene Parsing in the Wild}},
  author={Miao, Jiaxu and Wei, Yunchao and Wu, Yu and Liang, Chen and Li, Guangrui and Yang, Yi},
  booktitle={CVPR},
  year={2021}
}

@inproceedings{micheli2024delta,
  title={{Efficient World Models with Context-Aware Tokenization}},
  author={Micheli, Vincent and Alonso, Eloi and Fleuret, Fran{\c{c}}ois},
  booktitle={ICML},
  year={2024}
}

@misc{openai_sora_2024,
  title        = {{Sora}},
  author       = {{OpenAI}},
  year         = {2024},
  howpublished = {\url{https://openai.com/sora}}
}

@article{oquab2023dinov2,
  title={{DINOv2: Learning Robust Visual Features without Supervision}},
  author={Oquab, Maxime and Darcet, Timoth{\'e}e and Moutakanni, Th{\'e}o and Vo, Huy and Szafraniec, Marc and Khalidov, Vasil and Fernandez, Pierre and Haziza, Daniel and Massa, Francisco and El-Nouby, Alaaeldin and others},
  journal=TMLR,
  year={2024}
}

@inproceedings{peebles2023scalable,
  title={{Scalable Diffusion Models with Transformers}},
  author={Peebles, William and Xie, Saining},
  booktitle={ICCV},
  year={2023}
}

@inproceedings{podell2023sdxl,
  title={{SDXL: Improving Latent Diffusion Models for High-Resolution Image Synthesis}},
  author={Podell, Dustin and English, Zion and Lacey, Kyle and Blattmann, Andreas and Dockhorn, Tim and M{\"u}ller, Jonas and Penna, Joe and Rombach, Robin},
  booktitle={ICLR},
  year={2024}
}

@inproceedings{radford2021learning,
  title={{Learning Transferable Visual Models From Natural Language Supervision}},
  author={Radford, Alec and Kim, Jong Wook and Hallacy, Chris and Ramesh, Aditya and Goh, Gabriel and Agarwal, Sandhini and Sastry, Girish and Askell, Amanda and Mishkin, Pamela and Clark, Jack and others},
  booktitle={ICML},
  year={2021}
}

@inproceedings{rasley2020deepspeed,
  title={{DeepSpeed: System Optimizations Enable Training Deep Learning Models with Over 100 Billion Parameters}},
  author={Rasley, Jordan and Rajbhandari, Samyam and Ruwase, Olatunji and He, Yuxiong},
  booktitle={KDD},
  year={2020}
}

@inproceedings{ren2024flowar,
  title={{FlowAR: Scale-wise Autoregressive Image Generation Meets Flow Matching}},
  author={Ren, Sucheng and Yu, Qihang and He, Ju and Shen, Xiaohui and Yuille, Alan and Chen, Liang-Chieh},
  booktitle={ICML},
  year={2025}
}

@inproceedings{ren2025beyond,
  title={{Beyond Next-Token: Next-X Prediction for Autoregressive Visual Generation}},
  author={Ren, Sucheng and Yu, Qihang and He, Ju and Shen, Xiaohui and Yuille, Alan and Chen, Liang-Chieh},
  booktitle={ICCV},
  year={2025}
}

@article{ren2025grouping,
  title={{Grouping First, Attending Smartly: Training-Free Acceleration for Diffusion Transformers}},
  author={Ren, Sucheng and Yu, Qihang and He, Ju and Yuille, Alan and Chen, Liang-Chieh},
  journal={arXiv preprint arXiv:2505.14687},
  year={2025}
}

@inproceedings{rombach2022high,
  title={{High-Resolution Image Synthesis with Latent Diffusion Models}},
  author={Rombach, Robin and Blattmann, Andreas and Lorenz, Dominik and Esser, Patrick and Ommer, Bj{\"o}rn},
  booktitle={CVPR},
  year={2022}
}

@inproceedings{shin2025deeply,
  title={{Deeply Supervised Flow-Based Generative Models}},
  author={Shin, Inkyu and Yang, Chenglin and Chen, Liang-Chieh},
  booktitle={ICCV},
  year={2025}
}

@article{simeoni2025dinov3,
  title={{DINOv3}},
  author={Sim{\'e}oni, Oriane and Vo, Huy V and Seitzer, Maximilian and Baldassarre, Federico and Oquab, Maxime and Jose, Cijo and Khalidov, Vasil and Szafraniec, Marc and Yi, Seungeun and Ramamonjisoa, Micha{\"e}l and others},
  journal={arXiv preprint arXiv:2508.10104},
  year={2025}
}

@article{su2024roformer,
  title={{RoFormer: Enhanced Transformer with Rotary Position Embedding}},
  author={Su, Jianlin and Ahmed, Murtadha and Lu, Yu and Pan, Shengfeng and Bo, Wen and Liu, Yunfeng},
  journal={Neurocomputing},
  volume={568},
  pages={127063},
  year={2024},
}

@article{sun2024autoregressive,
  title={{Autoregressive Model Beats Diffusion: Llama for Scalable Image Generation}},
  author={Sun, Peize and Jiang, Yi and Chen, Shoufa and Zhang, Shilong and Peng, Bingyue and Luo, Ping and Yuan, Zehuan},
  journal={arXiv preprint arXiv:2406.06525},
  year={2024}
}

@inproceedings{teed2020raft,
  title={{RAFT: Recurrent All-Pairs Field Transforms for Optical Flow}},
  author={Teed, Zachary and Deng, Jia},
  booktitle=ECCV,
  year={2020}
}

@article{tian2024visual,
  title={{Visual Autoregressive Modeling: Scalable Image Generation via Next-Scale Prediction}},
  author={Tian, Keyu and Jiang, Yi and Yuan, Zehuan and Peng, Bingyue and Wang, Liwei},
  journal=NIPS,
  year={2024}
}

@inproceedings{touvron2021cait,
  title={{Going Deeper with Image Transformers}},
  author={Touvron, Hugo and Cord, Matthieu and Sablayrolles, Alexandre and Synnaeve, Gabriel and J{\'e}gou, Herv{\'e}},
  booktitle={ICCV},
  year={2021}
}

@article{tschannen2025siglip,
  title={{SigLIP 2: Multilingual Vision-Language Encoders with Improved Semantic Understanding, Localization, and Dense Features}},
  author={Tschannen, Michael and Gritsenko, Alexey and Wang, Xiao and Naeem, Muhammad Ferjad and Alabdulmohsin, Ibrahim and Parthasarathy, Nikhil and Evans, Talfan and Beyer, Lucas and Xia, Ye and Mustafa, Basil and others},
  journal={arXiv preprint arXiv:2502.14786},
  year={2025}
}

@article{van2017neural,
  title={{Neural Discrete Representation Learning}},
  author={Van Den Oord, Aaron and Vinyals, Oriol and others},
  journal=NIPS,
  year={2017}
}

@article{vaswani2017attention,
  title={{Attention is All You Need}},
  author={Vaswani, Ashish and Shazeer, Noam and Parmar, Niki and Uszkoreit, Jakob and Jones, Llion and Gomez, Aidan N and Kaiser, {\L}ukasz and Polosukhin, Illia},
  journal=NIPS,
  year={2017}
}

@inproceedings{vincent2008extracting,
  title={{Extracting and Composing Robust Features with Denoising Autoencoders}},
  author={Vincent, Pascal and Larochelle, Hugo and Bengio, Yoshua and Manzagol, Pierre-Antoine},
  booktitle={ICML},
  year={2008}
}

@inproceedings{walker2016uncertain,
  title={{An Uncertain Future: Forecasting from Static Images Using Variational Autoencoders}},
  author={Walker, Jacob and Doersch, Carl and Gupta, Abhinav},
  booktitle={ECCV},
  year={2016}
}

@article{walker2025generalist,
  title        = {{Generalist Forecasting with Frozen Video Models via Latent Diffusion}},
  author       = {Walker, Jacob C. and Vélez, Pedro and Polania Cabrera, Luisa and Zhou, Guangyao and Kabra, Rishabh and Doersch, Carl and Ovsjanikov, Maks and Carreira, João and Ginosar, Shiry},
  year         = {2025},
  journal      = {arXiv preprint arXiv:2507.13942},
}

@article{weber2024maskbit,
  title={{MaskBit: Embedding-free Image Generation via Bit Tokens}},
  author={Weber, Mark and Yu, Lijun and Yu, Qihang and Deng, Xueqing and Shen, Xiaohui and Cremers, Daniel and Chen, Liang-Chieh},
  journal=TMLR,
  year={2024}
}

@article{wiegand2003overview,
  title={{Overview of the H.264/AVC Video Coding Standard}},
  author={Wiegand, Thomas and Sullivan, Gary J and Bjontegaard, Gisle and Luthra, Ajay},
  journal={IEEE Transactions on circuits and systems for video technology},
  volume={13},
  number={7},
  pages={560--576},
  year={2003}
}

@article{williams1989learning,
  title={{A Learning Algorithm for Continually Running Fully Recurrent Neural Networks}},
  author={Williams, Ronald J and Zipser, David},
  journal={Neural computation},
  volume={1},
  number={2},
  pages={270--280},
  year={1989},
}

@inproceedings{wolf-etal-2020-transformers,
    title = {{Transformers: State-of-the-Art Natural Language Processing}},
    author = {Thomas Wolf and Lysandre Debut and Victor Sanh and Julien Chaumond and Clement Delangue and Anthony Moi and Pierric Cistac and Tim Rault and Rémi Louf and Morgan Funtowicz and Joe Davison and Sam Shleifer and Patrick von Platen and Clara Ma and Yacine Jernite and Julien Plu and Canwen Xu and Teven Le Scao and Sylvain Gugger and Mariama Drame and Quentin Lhoest and Alexander M. Rush},
    booktitle = {EMNLP Demos},
    year = {2020},
}

@inproceedings{xu2020vpeg,
  title     = {{Video Prediction via Example Guidance}},
  author    = {Xu, Jingwei and Xu, Huazhe and Ni, Bingbing and Yang, Xiaokang and Darrell, Trevor},
  booktitle = {ICML},
  year      = {2020}
}

@article{yang20241,
  title={{1.58-bit FLUX}},
  author={Yang, Chenglin and Liu, Celong and Deng, Xueqing and Kim, Dongwon and Mei, Xing and Shen, Xiaohui and Chen, Liang-Chieh},
  journal={arXiv preprint arXiv:2412.18653},
  year={2024}
}

@inproceedings{yao2025reconstruction,
  title={{Reconstruction vs. Generation: Taming Optimization Dilemma in Latent Diffusion Models}},
  author={Yao, Jingfeng and Yang, Bin and Wang, Xinggang},
  booktitle={CVPR},
  year={2025}
}

@inproceedings{yu2020bdd100k,
  title={{BDD100K: A Diverse Driving Dataset for Heterogeneous Multitask Learning}},
  author={Yu, Fisher and Chen, Haofeng and Wang, Xin and Xian, Wenqi and Chen, Yingying and Liu, Fangchen and Madhavan, Vashisht and Darrell, Trevor},
  booktitle={CVPR},
  year={2020}
}

@article{yu2022coca,
  title={{CoCa: Contrastive Captioners are Image-Text Foundation Models}},
  author={Yu, Jiahui and Wang, Zirui and Vasudevan, Vijay and Yeung, Legg and Seyedhosseini, Mojtaba and Wu, Yonghui},
  journal=TMLR,
  year={2022}
}

@article{yu2022scaling,
  title={{Scaling Autoregressive Models for Content-Rich Text-to-Image Generation}},
  author={Yu, Jiahui and Xu, Yuanzhong and Koh, Jing Yu and Luong, Thang and Baid, Gunjan and Wang, Zirui and Vasudevan, Vijay and Ku, Alexander and Yang, Yinfei and Ayan, Burcu Karagol and others},
  journal=TMLR,
  year={2022}
}

@inproceedings{yu2023language,
  title={{Language Model Beats Diffusion--Tokenizer is Key to Visual Generation}},
  author={Yu, Lijun and Lezama, Jos{\'e} and Gundavarapu, Nitesh B and Versari, Luca and Sohn, Kihyuk and Minnen, David and Cheng, Yong and Birodkar, Vighnesh and Gupta, Agrim and Gu, Xiuye and others},
  booktitle={ICLR},
  year={2024}
}

@article{yu2024image,
  title={{An Image is Worth 32 Tokens for Reconstruction and Generation}},
  author={Yu, Qihang and Weber, Mark and Deng, Xueqing and Shen, Xiaohui and Cremers, Daniel and Chen, Liang-Chieh},
  journal=NIPS,
  year={2024}
}

@inproceedings{yu2025randomized,
  title={{Randomized Autoregressive Visual Generation}},
  author={Yu, Qihang and He, Ju and Deng, Xueqing and Shen, Xiaohui and Chen, Liang-Chieh},
  booktitle={ICCV},
  year={2025}
}

@article{yu2026autoregressive,
  title={{Autoregressive Image Generation with Masked Bit Modeling}},
  author={Yu, Qihang and Liu, Qihao and He, Ju and Zhang, Xinyang and Liu, Yang and Chen, Liang-Chieh and Chen, Xi},
  journal={arXiv preprint arXiv:2602.09024},
  year={2026}
}

@inproceedings{zhai2023sigmoid,
  title={{Sigmoid Loss for Language Image Pre-Training}},
  author={Zhai, Xiaohua and Mustafa, Basil and Kolesnikov, Alexander and Beyer, Lucas},
  booktitle={ICCV},
  year={2023}
}

@inproceedings{zheng2025diffusiontransformersrepresentationautoencoders,
  title={{Diffusion Transformers with Representation Autoencoders}},
  author={Boyang Zheng and Nanye Ma and Shengbang Tong and Saining Xie},
  booktitle={ICLR},
  year={2026},
}

@inproceedings{zhou2024dino,
  title={{DINO-WM: World Models on Pre-trained Visual Features enable Zero-shot Planning}},
  author={Zhou, Gaoyue and Pan, Hengkai and LeCun, Yann and Pinto, Lerrel},
  booktitle={ICML},
  year={2025}
}
}
\clearpage
\renewcommand{\thetable}{\Alph{table}}
\setcounter{table}{0}
\renewcommand{\thefigure}{\Alph{figure}}
\setcounter{figure}{0}

\appendix
\crefname{appendix}{Appendix}{Appendices}
\Crefname{appendix}{Appendix}{Appendices}
\crefalias{section}{appendix}
\section*{Appendix}
\label{appendix}
\paragraph{Table of contents:}
\begin{itemize}[itemsep=-1pt,topsep=-1pt]
\item \Cref{sec:supp:impl_details}: Additional Implementation Details
\item \Cref{sec:supp:eval_details}: Additional Evaluation Details
\item \Cref{sec:supp:disc_abl}: Delta Tokens in Discriminative Models
\item \Cref{sec:supp:limitations}: Limitations and Future Work
\item \Cref{sec:supp:qualitative}: Additional Qualitative Examples
\end{itemize}

\section{Additional Implementation Details}
\label{sec:supp:impl_details}

\paragraph{DeltaTok tokenizer.}
Our \textit{\tokenizername} tokenizer is a simple continuous auto-encoder~\cite{hinton2006reducing}, not a variational auto-encoder (VAE)~\cite{kingma2013auto}. It compresses the patch tokens from the DINOv3~\cite{simeoni2025dinov3} ViT-B~\cite{dosovitskiy2021image} VFM, which uses a patch size of $16\times16$. Both the tokenizer encoder and decoder use the ViT-B configuration, reusing the DINOv3 Transformer block implementation from Hugging Face Transformers~\cite{wolf-etal-2020-transformers}, including 2D RoPE for spatial position encoding, but skipping the patch embedding layer because the tokenizer operates on VFM output patch tokens rather than pixels. The encoder adds a learned per-frame embedding to each input token, distinguishing previous-frame from current-frame tokens. All linear and embedding weights are initialized with truncated normal ($\sigma{=}0.02$), linear biases are set to zero, and Layer Scale~\cite{touvron2021cait} values are initialized to $10^{-5}$. In the tokenizer decoder, we omit the final layer normalization so that the small initial Layer Scale values make the decoder behave approximately as an identity map at initialization.

\paragraph{Tokenizer training.}
We train the tokenizer on sampled frame pairs for 50K iterations with a mean squared error (MSE) loss, using AdamW~\cite{loshchilov2019adamw} with linear warmup to $10^{-3}$ over 5K steps and a constant learning rate thereafter, weight decay of $10^{-4}$, a batch size of $1{,}024$, and gradient norm clipping at $10^{-2}$.

\paragraph{DINO-world predictor reimplementation.}
An official DINO-world codebase has not been released, so all DINO-world baselines in this paper use our own reimplementation following the protocol described in DINO-world~\cite{baldassarre2025back}. We use the ViT-B configuration for the predictor. Specifically, spatial and temporal identity are injected through axial rotary positional embeddings (3D RoPE~\cite{su2024roformer}) applied to the query and key projections, rotating the first $20{+}20{+}20$ dimensions per head and leaving the final $4$ unrotated. Furthermore, spatial predictions of frame $t{+}1$ are computed using a block-causal attention mask during training, ensuring queries only attend to past frames while allowing efficient parallelization. Weight initialization follows the tokenizer (see above).

\paragraph{DeltaWorld predictor.}
The future predictor also uses the ViT-B configuration.
Because each frame is represented by a single token rather than an $H \times W$ grid, neither the block-causal attention mask nor the three\mbox{-}dimensional RoPE used in DINO-world is needed. We therefore simplify the block-causal mask to a standard causal (diagonal) mask, and the 3D RoPE to a 1D variant that rotates the first $60$ dimensions of each head, again leaving the final $4$ unrotated. Noise queries are sampled from $\mathcal{N}(0,\,0.02^{2} I)$. Weight initialization follows the tokenizer (see above).

\paragraph{Predictor training.}
The DINO-world and \modelname predictors share the same training configuration~\cite{baldassarre2025back}: AdamW~\cite{loshchilov2019adamw}, a learning rate of $10^{-4}$ with linear warmup over 5K steps and a constant learning rate thereafter, weight decay $4{\times}10^{-1}$, smooth L1 loss with $\beta{=}0.1$, a batch size of $1{,}024$, a training sequence length of $8$ frames, and no gradient clipping. For the main results, predictors are trained for 300K iterations; for ablations, this is reduced to 100K. The predictors are subsequently fine-tuned for 5K iterations at a $10\times$ lower learning rate.

\paragraph{Training augmentations.}
For all models (tokenizers and predictors), we use random resized crops with a scale range of 0.6--1.0 and an aspect-ratio range of 3{:}4--4{:}3 applied to the original frames. The resulting crop coordinates are applied consistently to all frames in the sequence, and the crop is then resized to a square, introducing a small amount of aspect-ratio distortion. Temporal offsets $\Delta\tau$ between consecutive frames are sampled uniformly from $[1/25,\,1/3]$ seconds.

\paragraph{Training data statistics.}
\begin{table}[t!]
\centering
\small
\renewcommand{\arraystretch}{1.}
\setlength{\tabcolsep}{3pt}
\begin{tabular*}{\linewidth}{@{\extracolsep{\fill}} l c c c}
\toprule
                 & Num. samples & Duration (s) & FPS \\
\midrule
DINO-world~\cite{baldassarre2025back}
                 & ${\sim}$66M & 5--60 & 10--60 \\
Ours
                 & ${\sim}$4M & 11 & 16 \\
\bottomrule
\end{tabular*}
\caption{\textbf{Training data statistics.}
For DINO-world~\cite{baldassarre2025back}, we report the duration range and FPS from their paper. For ours, we report the mean duration, and all videos have the same frame rate.}
\label{tab:train-dataset-stats}
\end{table}

Similar to the experimental setting of DINO-world~\cite{baldassarre2025back}, all models (tokenizers and predictors) are trained on a large collection of videos spanning diverse domains. The training data used for DINO-world is not publicly released; \Cref{tab:train-dataset-stats} compares ours with what is reported in DINO-world. Our dataset comprises videos mostly at $640{\times}360$ resolution, spanning a wide range of scenarios similar in spirit to the DINO-world corpus.

\paragraph{Task heads.}
Following DINO-world~\cite{baldassarre2025back}, linear segmentation and depth heads are trained on frozen VFM features from the training split of each evaluation dataset. For segmentation on VSPW~\cite{miao2021vspw} and Cityscapes~\cite{cordts2016cityscapes}, the head uses a batch normalization layer followed by a linear layer projecting to 124 and 19 semantic classes, respectively. For depth estimation on KITTI~\cite{geiger2013vision}, we follow the DINOv3~\cite{simeoni2025dinov3} depth head architecture. Specifically, a batch normalization layer and a linear layer produce 256 logits per pixel. These logits are rectified and shifted by $\epsilon=0.1$, normalized across the 256 bins to form a discrete depth distribution, and then mapped to a continuous depth by taking the expectation over 256 uniformly spaced bins between $10^{-3}$ and $80$\,m. Depth evaluation is restricted to valid pixels within the Garg region~\cite{garg2016unsupervised}.

\section{Additional Evaluation Details}
\label{sec:supp:eval_details}

\paragraph{Sequences.}
We extract evaluation sequences following DINO-world~\cite{baldassarre2025back}. We use the validation split for VSPW~\cite{miao2021vspw} and Cityscapes~\cite{cordts2016cityscapes}, and the Eigen test split~\cite{eigen2014depth} for KITTI~\cite{geiger2013vision}. Time strides are 0.2\,s for VSPW and KITTI, and 0.1875\,s for Cityscapes. For VSPW, we select every 20th frame for evaluation and extract non-overlapping subsequences to keep the total number of sequences manageable.

\paragraph{Evaluation pre- and postprocessing.}
Training uses square inputs, while evaluation datasets contain rectangular images.
Therefore, during evaluation, frames are resized so that the shorter side matches the input size used in each experiment (512 in the main setting and 256 in the ablation setting). For KITTI, the Eigen crop ($352{\times}1216$) is applied to frames and depth maps before resizing. After cropping, frames are squashed to a 1{:}2 aspect ratio for fair comparison with Cosmos~\cite{agarwal2025cosmos}, whose input format does not support wider frames. We then take two potentially overlapping left/right square crops from the resized frames. Labels are not resized, but split at the horizontal midpoint into two non-overlapping halves that define the regions used for evaluation. After generating future features, task outputs from each crop are bilinearly upsampled and cropped to match the corresponding label half for evaluation.

\paragraph{Cosmos.}
Cosmos (Predict1)~\cite{agarwal2025cosmos} can only be evaluated under its native inference constraints, and we follow a similar protocol to DINO-world~\cite{baldassarre2025back}. 
Specifically, Cosmos requires a fixed context of $9$ input frames and generates a rollout of $24$ future frames in a single forward pass. Frames are resized so that the height is $512$ pixels while preserving the aspect ratio, and padded to $640\times1024$ as required by the Cosmos input format.
For KITTI, the Eigen crop is applied before resizing to $512\times1024$, which squashes the aspect ratio to $1{:}2$. For all other datasets, no cropping is applied before generation. After generation, we remove the padding and apply the same left/right cropping protocol as above before re-encoding each predicted crop with DINOv3, ensuring consistent evaluation with other models.

\paragraph{\textit{Best} and \textit{mean} evaluation.}
We generate 20 independent rollouts per sequence, unless noted otherwise.
The \textit{best} score is computed on the rollout whose DINOv3
features have the lowest feature-space loss to the ground truth at the last predicted timestep.
The \textit{mean} score averages the 20 DINOv3 features at the last predicted timestep and then applies
the task head once on the averaged features. We do not average scores from individual predictions, as averaging in feature space enables fair comparison with discriminative models that produce a single prediction. This evaluation protocol is applied per crop, identically to \modelname and Cosmos. For the discriminative DINO-world baseline, we report the score of its single deterministic prediction.

\paragraph{FLOPs.}
All GFLOPs are computed for square inputs and doubled, since evaluation uses two square-crop forward passes as described above. Cosmos is the exception, as it does not use square crops.
Additionally, for Cosmos we exclude the fixed-cost GFLOPs associated with the tokenizer and KV pre-filling, which we expect to be small relative to the autoregressive decoding and iterative diffusion.
For step~\tablestep{2} in \Cref{tab:steps-results}, GFLOPs include applying the tokenizer decoder at each intermediate rollout step, not only the final one.

\paragraph{Training time and memory.} In \Cref{tab:steps-results}, we measure the training time per optimization iteration and steady-state GPU memory on a single node with 8 NVIDIA H200 GPUs, using BF16 mixed precision and \texttt{torch.compile} (default mode). Despite generating $K{=}16$ candidate futures, BoM training in step~\tablestep{1} requires similar memory to the discriminative baseline, because the candidate selection pass uses detached parameters (no activation storage for backpropagation) and only the best candidate is re-run with gradients. Delta compression in step~\tablestep{3} is slightly slower than frame compression in step~\tablestep{2} because its tokenizer encoder processes both the current and previous frame's patch tokens.

\paragraph{Efficiency breakdown.}
\begin{table}[t!]
\centering
\small
\renewcommand{\arraystretch}{1.}
\setlength{\tabcolsep}{6pt}
\begin{tabular*}{\linewidth}{@{\extracolsep{\fill}} l r}
\toprule
\multicolumn{2}{c}{\textbf{Discriminative DINO-world~\cite{baldassarre2025back}}} \\
\midrule
{\color{gray}Backbone (4 frames)} & {\color{gray}$4\times47.185=188.74$} \\
Predictor (4-frame context) & 84.88 \\
Predictor (5-frame context) & 96.96 \\
Predictor (6-frame context) & 109.04 \\
\midrule
\\[-4pt]
\multicolumn{2}{c}{\textbf{Generative DeltaWorld (Ours)}} \\
\midrule
\multicolumn{2}{@{}l}{\textit{Shared once}} \\
{\color{gray}Backbone (4 frames)} & {\color{gray}$4\times47.185=188.74$} \\
{\color{gray}DeltaTok encoder (4 frames)} & {\color{gray}$4\times96.930=387.72$} \\
\midrule
\multicolumn{2}{@{}l}{\textit{Per sample (repeated $K$ times)}} \\
Predictor (4-frame context) & 0.26 \\
Predictor (5-frame context) & 0.28 \\
Predictor (6-frame context) & 0.31 \\
DeltaTok decoder (step 1) & 46.12 \\
DeltaTok decoder (step 2) & 46.12 \\
DeltaTok decoder (step 3) & 46.12 \\
\bottomrule
\end{tabular*}
\caption{\textbf{GFLOPs breakdown.} In \modelname, the backbone and \tokenizername encoder run once, while the predictor and \tokenizername decoder are applied per generated sample. Using a three-step rollout and a four-frame context (mid-horizon), ViT-B components, and $256\times256$ crops.}
\label{tab:supp:efficiency}
\end{table}

\Cref{tab:supp:efficiency} shows how GFLOPs are distributed across the model components for both the discriminative DINO-world~\cite{baldassarre2025back} and our generative \modelname. Although the predictor dominates compute in DINO-world, its cost becomes negligible in \modelname with a short context of four to six delta tokens, with most per-sample compute instead coming from the \tokenizername decoder. Crucially, however, unlike the predictor in DINO-world, the decoder’s compute cost does not increase with context length. Even with the small predictor size and the benchmark’s short context length, the decoder remains more efficient than the predictor in DINO-world. Furthermore, the \tokenizername encoder overhead in \modelname is shared across all generated samples. This makes \modelname noticeably cheaper per generated sample and enables efficient multi-sample generation, while the future predictor remains lightweight and flexible for scaling, \eg, in context or predictor size.

\section{Delta Tokens in Discriminative Models}
\label{sec:supp:disc_abl}
Although not the primary focus of this paper, \emph{delta tokens} can also be used in a discriminative world model.
\Cref{tab:delta_ablation} shows that replacing per-frame patch tokens with a single delta token in the discriminative DINO-world baseline~\cite{baldassarre2025back} performs well (-0.2 on VSPW and +1.5 on Cityscapes), while also being more efficient in training time (0.5$\times$) and memory (0.2$\times$).

\begin{table}[t!]
\centering
\small
\renewcommand{\arraystretch}{1.}
\setlength{\tabcolsep}{3pt}
\begin{tabular*}{\linewidth}{@{\extracolsep{\fill}} l r r r r}
\toprule
\textbf{Model}
& \textbf{Time}
& \textbf{Mem}
& \textbf{VSPW $\uparrow$}
& \textbf{Cityscapes $\uparrow$} \\
\midrule
\textit{Copy last (lower bound)}
  & \textit{--}
  & \textit{--}
  & \textit{41.8}
  & \textit{37.9} \\
\midrule
{DINO-world$^\dagger$~\cite{baldassarre2025back}}
  & 1.0$\times$
  & 1.0$\times$
  & \textbf{44.8}
  & 45.4 \\
\downrightarrow \textbf{Delta compression}
  & \textbf{0.5$\times$}
  & \textbf{0.2$\times$}
  & 44.6
  & \textbf{46.9} \\
\midrule
\textit{Present (upper bound)}
  & \textit{--}
  & \textit{--}
  & \textit{52.0}
  & \textit{59.3} \\
\bottomrule
\end{tabular*}
\caption{\textbf{Delta tokens in the discriminative DINO-world~\cite{baldassarre2025back}.} \emph{Delta tokens} also perform well within a discriminative world model. Time and Mem report per-iteration training time and GPU memory relative to the discriminative baseline. Reporting mid-horizon (${\sim}0.6$\,s) mIoU using $256\times256$ crops. $^\dagger$Our reimplementation.}
\label{tab:delta_ablation}
\end{table}

We also integrate delta tokens into DINO-Foresight~\cite{karypidis2024dino}, a separate discriminative world model with a different architecture, using their official implementation. It is trained and evaluated on Cityscapes~\cite{cordts2016cityscapes} and extracts multi-layer DINOv2~\cite{oquab2023dinov2} features, applying PCA to obtain 1152-dimensional spatial features per patch. We train a \tokenizername variant that compresses these PCA features of two consecutive frames into a single 1152-dimensional delta token at $448{\times}896$ resolution, using BDD100K~\cite{yu2020bdd100k} and briefly fine-tuning on Cityscapes~\cite{cordts2016cityscapes}. We then retrain the DINO-Foresight world model on Cityscapes to predict these delta tokens instead of spatial PCA features. Since delta tokens collapse the large spatio-temporal sequence to only one token per frame, we can simplify the architecture by replacing the factorized space-time attention with standard self-attention, and skip the high-resolution fine-tuning stage, training directly at the target resolution. As shown in \Cref{tab:foresight}, the delta-compressed variant matches the original while reducing the token count by $2048{\times}$, confirming that delta tokens transfer effectively across discriminative world model architectures.

\begin{table}[t!]
\centering
\small
\renewcommand{\arraystretch}{1.}
\setlength{\tabcolsep}{3.0pt}
\begin{tabular}{l c c c c c}
\toprule
& & \multicolumn{2}{c}{\textbf{Seg.\ mIoU $\uparrow$}}
& \multicolumn{2}{c}{\textbf{Depth $\delta_1$ $\uparrow$}} \\
\cmidrule(lr){3-4} \cmidrule(lr){5-6}
& \textbf{Tokens} & Short & Mid & Short & Mid \\
\midrule
\textit{Copy last (lower bound)}
&  & \textit{54.7} & \textit{40.4}
& \textit{84.1} & \textit{77.8} \\
\midrule
DINO-Foresight$^\dagger$~\cite{karypidis2024dino}
& 10240 & 71.8 & 59.8
& \textbf{88.6} & 85.4 \\
\downrightarrow \textbf{Delta compression}
& \textbf{5} & \textbf{72.1} & \textbf{60.0}
& 88.5 & \textbf{85.6} \\
\midrule
\textit{Present (upper bound)}
&  & \textit{77.0} & \textit{77.0}
& \textit{89.1} & \textit{89.1} \\
\bottomrule
\end{tabular}
\caption{\textbf{Delta tokens in the discriminative DINO-Foresight~\cite{karypidis2024dino}.} Results on Cityscapes~\cite{cordts2016cityscapes} show that \emph{delta tokens} transfer effectively to a different discriminative architecture, matching performance with $2048{\times}$ fewer tokens. The token count indicates the total number of tokens used by the world model. Using $448{\times}896$ frames. $^\dagger$Numbers reported in the DINO-Foresight paper~\cite{karypidis2024dino}.}
\label{tab:foresight}
\end{table}

\section{Limitations and Future Work}
\label{sec:supp:limitations}

We discuss two limitations of our work and directions for future research.

\paragraph{Distribution modeling.} The Best-of-Many (BoM)~\cite{bhattacharyya2018bom} objective enables efficient, non-iterative generation of diverse futures by mapping stochastic noise queries to distinct futures. However, unlike diffusion models~\cite{ho2020denoising}, whose denoising objective provides a principled connection to the data distribution, BoM lacks an explicit distributional objective. Consequently, the model’s coverage of the predictive distribution is limited by the number of noise queries $K$ explored during training, with no mechanism encouraging diverse utilization of the query space, and no guarantee that the distribution over sampled futures approximates the true probability of each outcome. That said, in practice different queries tend to produce distinct futures, suggesting the query space may serve as a form of implicit action conditioning. This could open a path toward explicit action-conditional generation, as similar queries may produce similar futures across different scenes.

\paragraph{Error accumulation.} Because delta tokens encode temporal differences, reconstructing absolute feature maps requires repeatedly decoding delta tokens conditioned on previous features. During tokenizer reconstruction, errors may compound across steps, potentially leading to feature drift. A natural mitigation is to have the tokenizer operate on its own reconstructions, computing delta tokens sequentially relative to previously decoded frames, rather than in parallel from ground truth input frames. In \modelname, the predictor may introduce an additional source of error, which may further compound during multi-step autoregressive rollouts, a well-known challenge in autoregressive video generation. Existing approaches to mitigate error accumulation in autoregressive generation may apply.

\section{Additional Qualitative Examples}
\label{sec:supp:qualitative}

\begin{figure*}
    \centering
    \includegraphics[width=\linewidth]{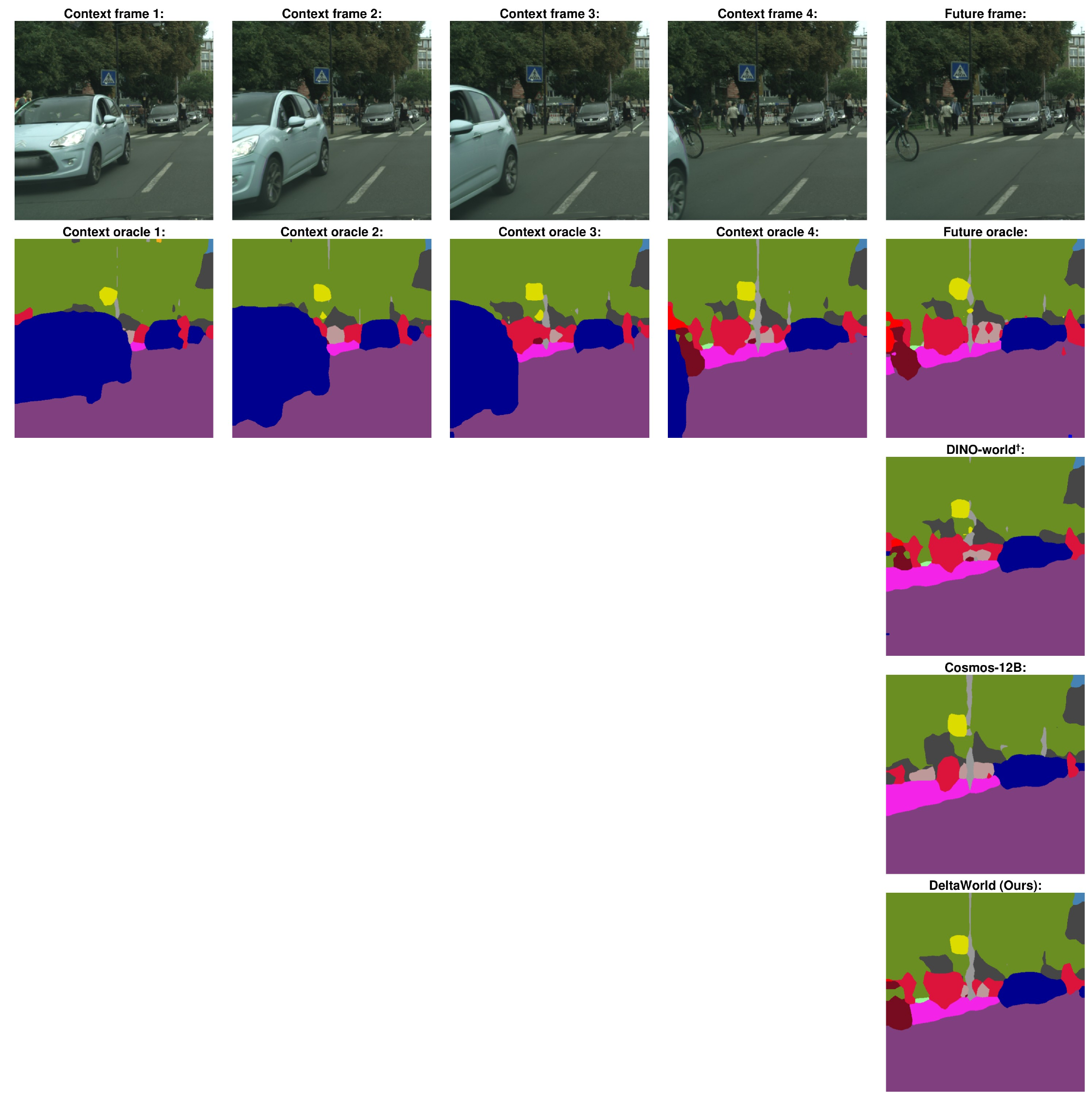}
    \caption{\textbf{DINO-world$^\dagger$~\cite{baldassarre2025back} \vs Cosmos-12B~\cite{agarwal2025cosmos} \vs \modelname (Ours).}
    Given a context of four frames, predict the fifth frame (short-horizon).
    Second row shows the segmentation head output on the ground-truth frames, while third, fourth, and fifth rows show the segmentation head output for the predicted future frame. In this Cityscapes example~\cite{cordts2016cityscapes}, \modelname provides an accurate prediction of the scene evolution.
    Generative models show \textit{mean} features from 20 samples; DINO-world shows its single deterministic prediction. Using $512\times512$ crops.
    $^\dagger$Our reimplementation.}
    \label{fig:cs_5_short}
\end{figure*}

In \Cref{fig:cs_5_short} we show short-horizon Cityscapes~\cite{cordts2016cityscapes} predictions from
DINO-world~\cite{baldassarre2025back}, Cosmos-12B~\cite{agarwal2025cosmos},
and \modelname. All three models predict the car moving out of the frame, but both DINO-world and Cosmos fail to maintain the bicycle wheel, DINO-world also loses the sign post, and Cosmos misses some of the people in the background.

\begin{figure*}
    \centering
    \includegraphics[width=\linewidth]{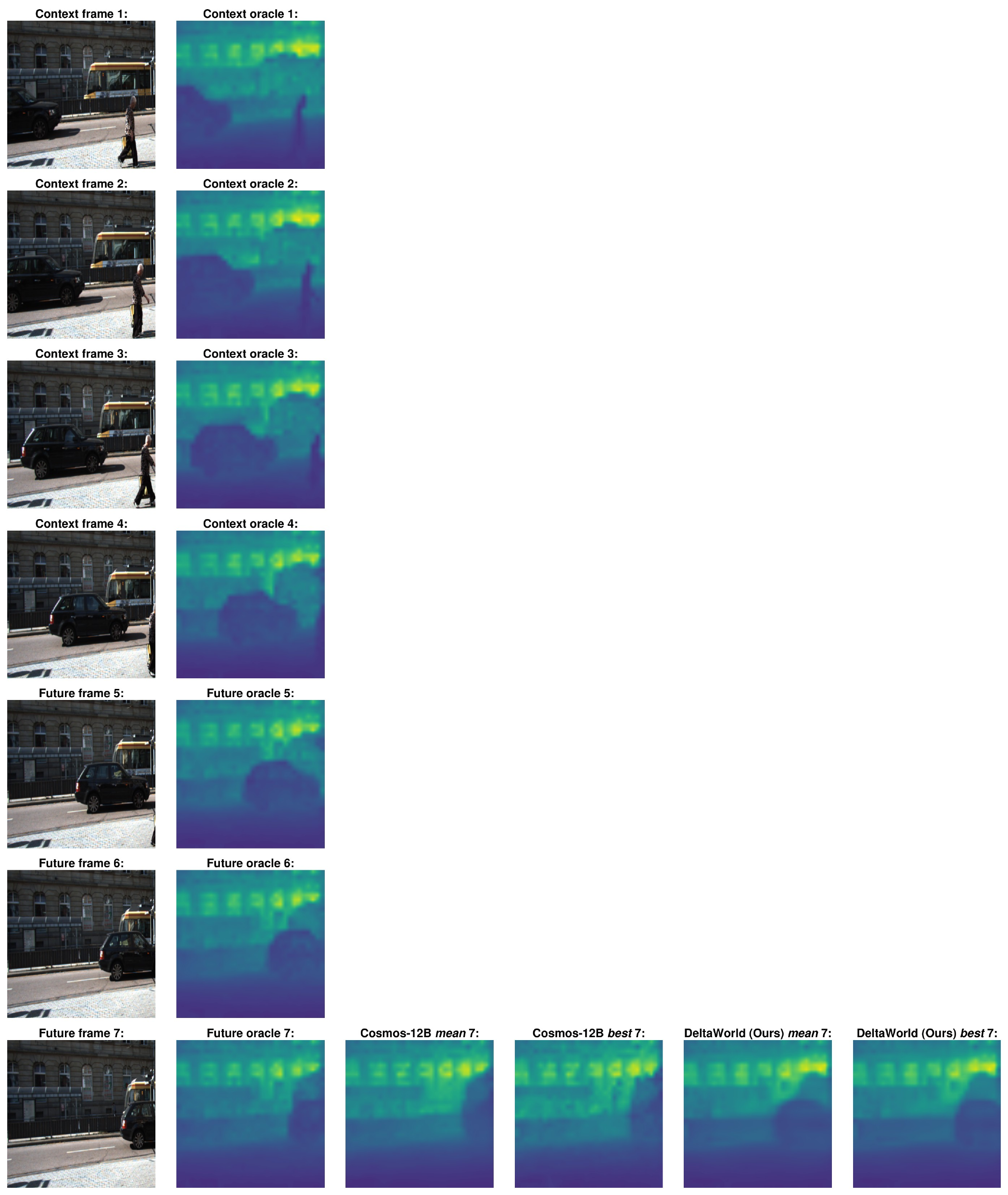}
    \caption{\textbf{Comparing \textit{mean} and \textit{best} for Cosmos-12B~\cite{agarwal2025cosmos} \vs \modelname (Ours).} Given a context of four frames, predict the seventh frame autoregressively (mid-horizon). Second column shows the depth head output on the ground-truth frames, third and fourth columns show Cosmos, and fifth and sixth columns show \modelname predictions. In this KITTI example~\cite{geiger2013vision}, \modelname's \textit{best} sample more closely matches the oracle depth layout. Using $512\times512$ crops.}
    \label{fig:kitti_14}
\end{figure*}

In \Cref{fig:kitti_14} we show mid-horizon KITTI~\cite{geiger2013vision} predictions, comparing
\textit{mean} and \textit{best} samples for Cosmos-12B and \modelname.
Both models track the car’s motion, but \modelname’s \textit{best} sample is more accurate than Cosmos’s: for example, it provides a more accurate depth estimate on the passing train.
Cosmos also yields \textit{mean} and \textit{best} samples that are
very similar, reflecting lower variation across its outputs.

\begin{figure*}
    \centering
    \includegraphics[width=0.90\linewidth]{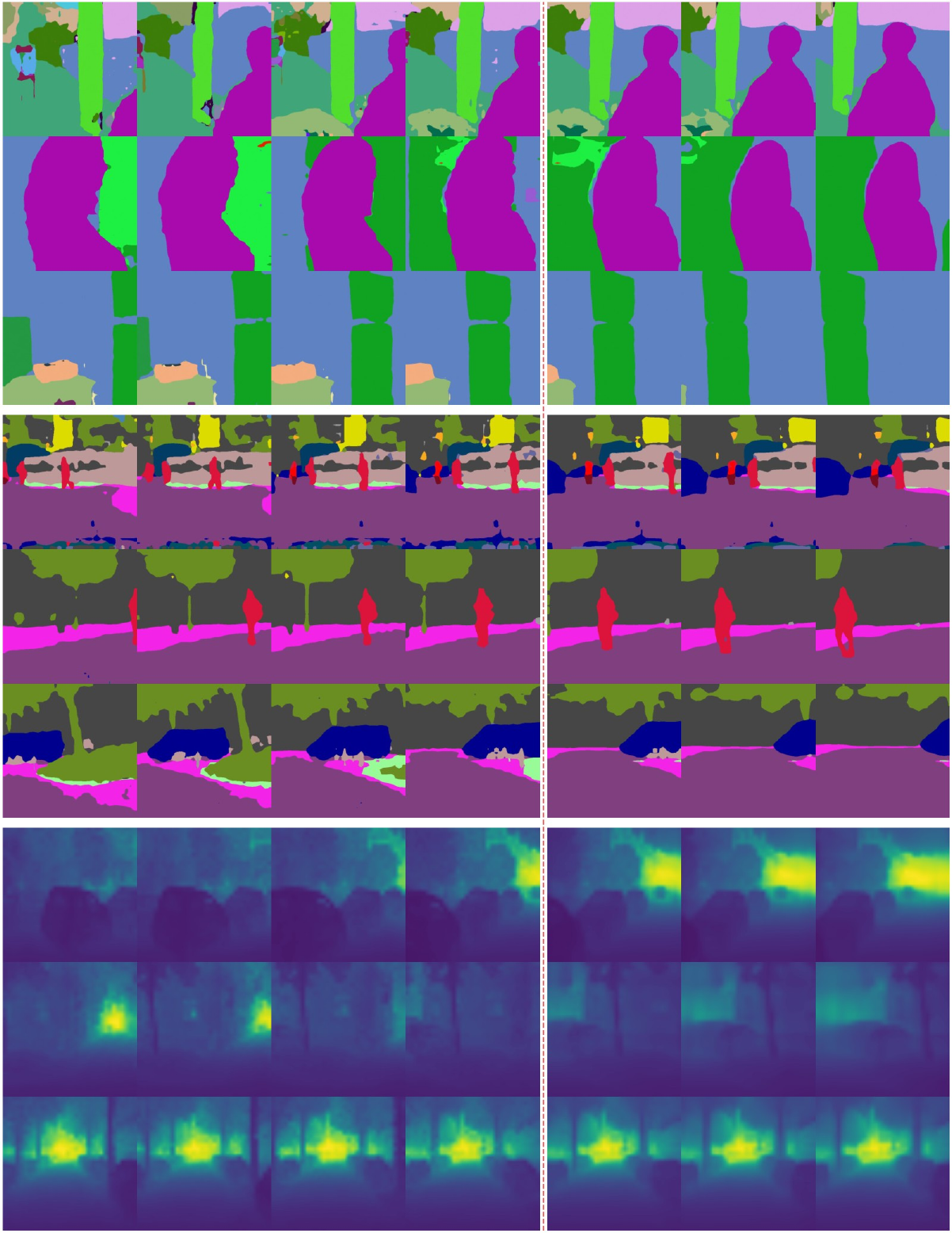}
    \caption{\textbf{Mid-horizon rollouts (task head visualization).}
    Each row shows four context frames (left of the dashed line) and an autoregressive rollout from \modelname (right), conditioned on random noise queries, in a single forward pass per step.
    Top: VSPW~\cite{miao2021vspw} segmentation, middle: Cityscapes~\cite{cordts2016cityscapes} segmentation, bottom: KITTI~\cite{geiger2013vision} depth.
    Using $512\times512$ crops.}
    \label{fig:multi_rollout_task}
\end{figure*}

\begin{figure*}
    \centering
    \includegraphics[width=0.90\linewidth]{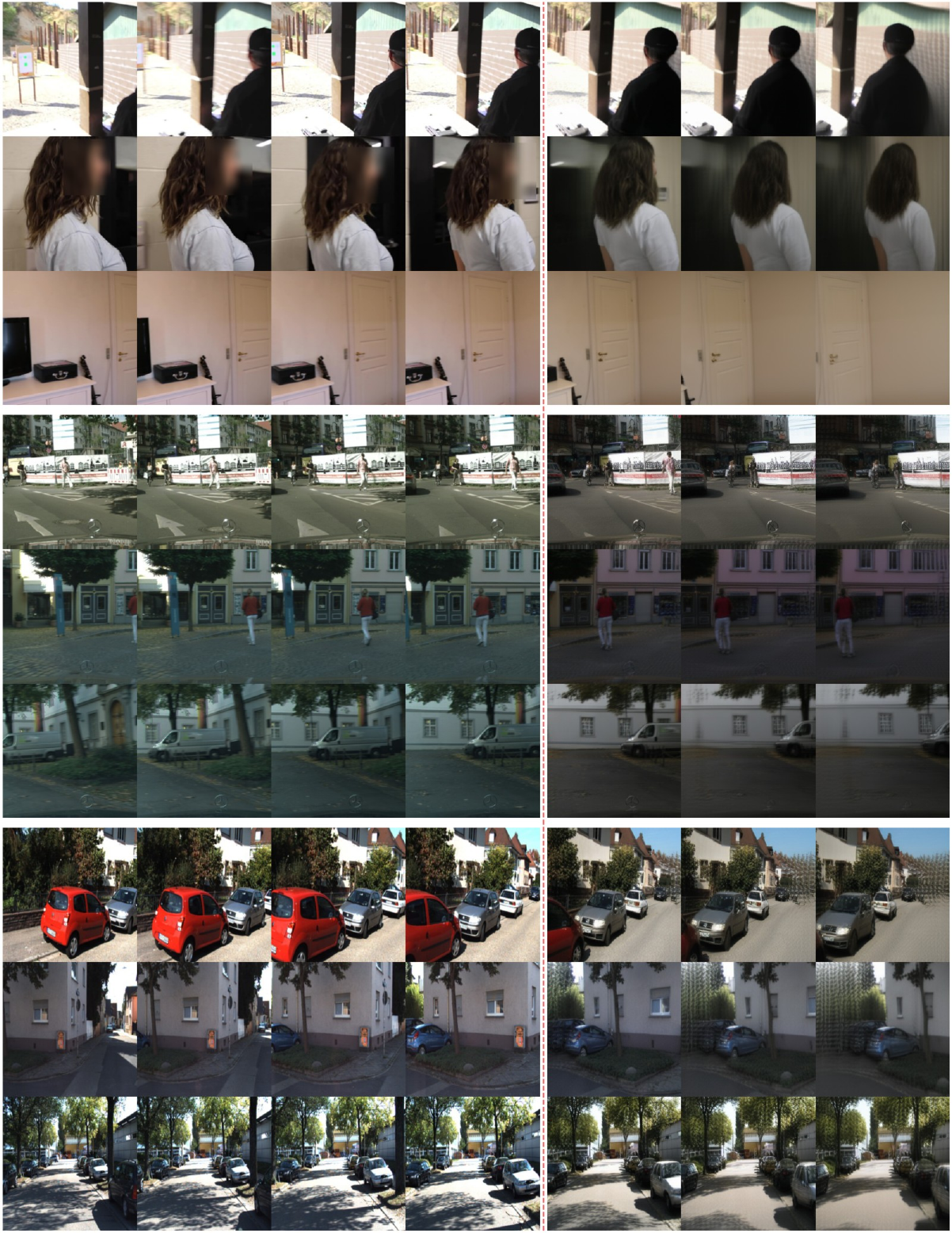}
    \caption{\textbf{Mid-horizon rollouts (RGB visualization).}
    Same sequences as~\Cref{fig:multi_rollout_task}.
    Context columns (left of the
dashed line) show ground-truth RGB frames; future columns show the predicted features decoded into pixels.
    Using $512\times512$ crops.}
    \label{fig:multi_rollout_rgb}
\end{figure*}

In \Cref{fig:multi_rollout_task,fig:multi_rollout_rgb} we show mid-horizon autoregressive rollouts from \modelname across all three evaluation datasets, visualized through task head outputs (\Cref{fig:multi_rollout_task}) and RGB reconstructions (\Cref{fig:multi_rollout_rgb}).
Since \modelname operates in DINOv3 feature space, we use the decoder from Representation Autoencoder (RAE)~\cite{zheng2025diffusiontransformersrepresentationautoencoders}, trained on ImageNet~\cite{deng2009imagenet} with DINOv3 ViT-B features, to decode predicted features back into pixels for the RGB visualization.

\begin{figure*}
    \centering
    \includegraphics[width=0.85\linewidth]{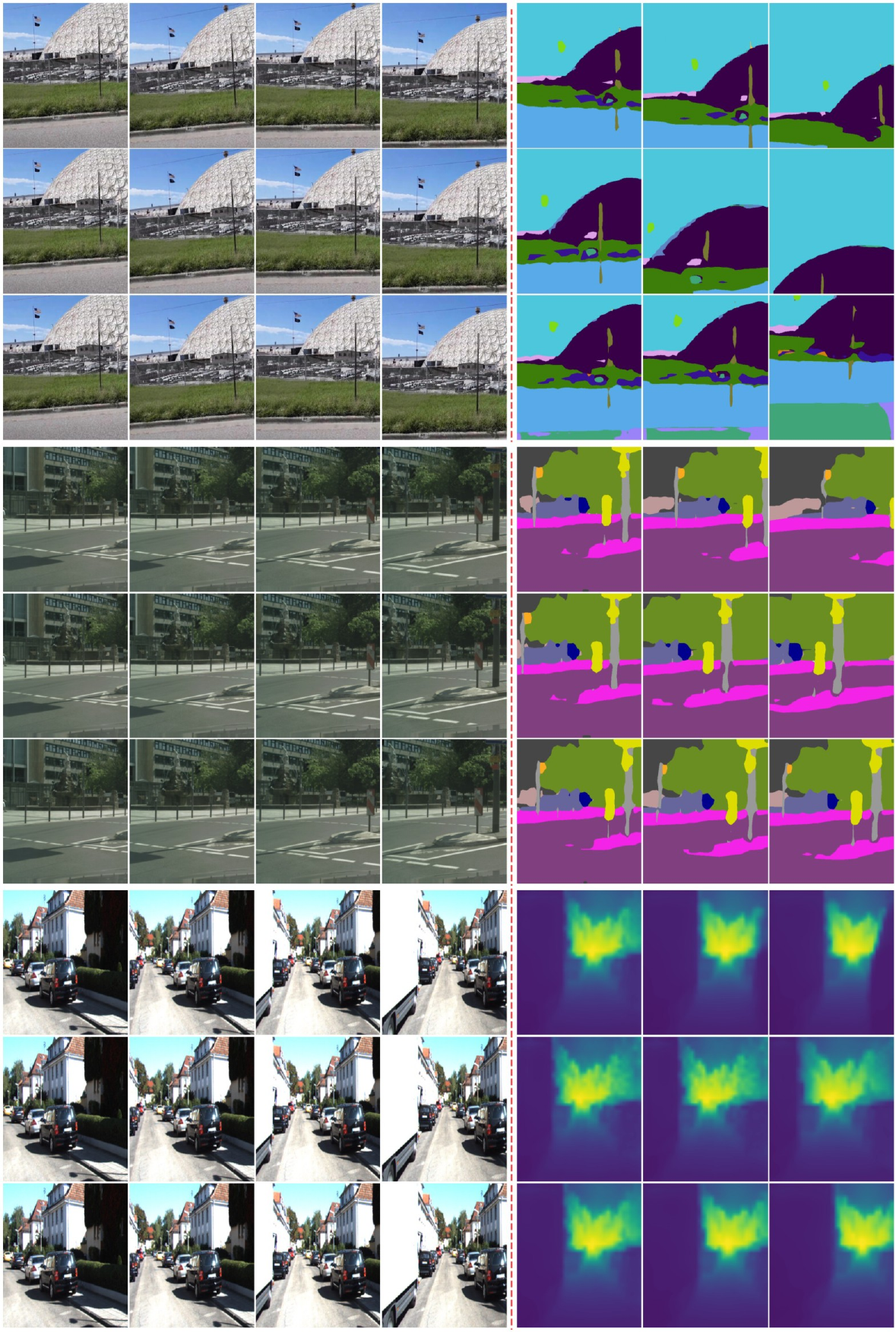}
    \caption{\textbf{Diverse mid-horizon rollouts (task head visualization).}
    Each group of three rows shares the same four context frames (left of the dashed line) but shows three different autoregressive rollouts from \modelname, each conditioned on random noise queries, in a single forward pass per step.
    Top: VSPW~\cite{miao2021vspw} segmentation, middle: Cityscapes~\cite{cordts2016cityscapes} segmentation, bottom: KITTI~\cite{geiger2013vision} depth.
    Using $512\times512$ crops.}
    \label{fig:diverse_rollout_task}
\end{figure*}

\begin{figure*}
    \centering
    \includegraphics[width=0.85\linewidth]{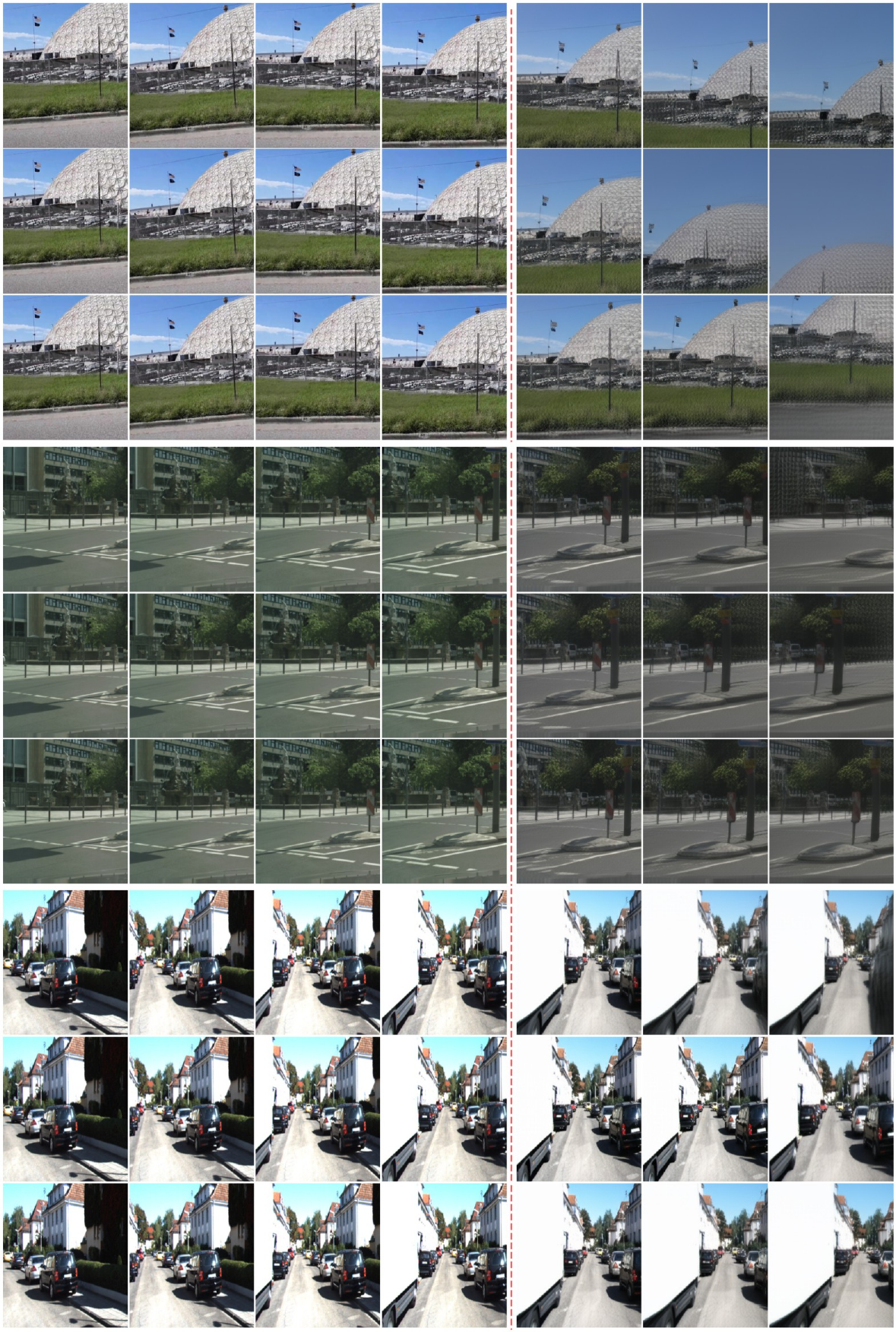}
    \caption{\textbf{Diverse mid-horizon rollouts (RGB visualization).}
    Same sequences and samples as~\Cref{fig:diverse_rollout_task}.
    Context columns (left of the
dashed line) show ground-truth RGB frames; future columns show the predicted features decoded into pixels.
    Using $512\times512$ crops.}
    \label{fig:diverse_rollout_rgb}
\end{figure*}

In \Cref{fig:diverse_rollout_task,fig:diverse_rollout_rgb} we visualize the diversity of mid-horizon autoregressive rollouts from \modelname across all three evaluation datasets by showing multiple samples for the same input context, again as task head outputs (\Cref{fig:diverse_rollout_task}) and RGB reconstructions (\Cref{fig:diverse_rollout_rgb}). Each group of three rows shares the same four context frames but shows three different rollouts.

\clearpage

\end{document}